\documentclass[10pt,twocolumn,letterpaper]{article}

\usepackage{cvpr}              

\usepackage{graphicx}
\usepackage{amsmath}
\usepackage{amssymb}
\usepackage{booktabs}

\usepackage{times}
\usepackage{amsmath}
\usepackage{amssymb}
\usepackage{siunitx}
\usepackage{cite}
\usepackage{ctable}
\usepackage{hhline}
\usepackage{booktabs} 
\usepackage{subcaption}
\usepackage{csquotes}
\usepackage{multirow}
\usepackage{pifont}
\usepackage{animate}

\usepackage{anyfontsize}
\usepackage[super]{nth}
\newcolumntype{L}[1]{>{\raggedright\let\newline\\\arraybackslash\hspace{0pt}}m{#1}}
\newcolumntype{C}[1]{>{\centering\let\newline\\\arraybackslash\hspace{0pt}}m{#1}}
\newcolumntype{R}[1]{>{\raggedleft\let\newline\\\arraybackslash\hspace{0pt}}m{#1}}

\usepackage{soul}
\usepackage{multicol}
\usepackage{makecell}
\usepackage{multirow}
\usepackage{booktabs}
\usepackage{color, colortbl}
\definecolor{Gray}{gray}{0.9}

\usepackage{paralist}
\usepackage{enumitem}

\usepackage[pagebackref,breaklinks,colorlinks]{hyperref}

\usepackage[capitalize]{cleveref}
\crefname{section}{Sec.}{Secs.}
\Crefname{section}{Section}{Sections}
\Crefname{table}{Table}{Tables}
\crefname{table}{Tab.}{Tabs.}

\def\cvprPaperID{5637} 
\def\confName{CVPR}
\def\confYear{2022}

\begin{document}

\title{Learning to Estimate Robust 3D Human Mesh from\\In-the-Wild Crowded Scenes}

\author{
Hongsuk Choi\hspace{1.0cm} Gyeongsik Moon\hspace{1.0cm} JoonKyu Park\hspace{1.0cm} Kyoung Mu Lee\\
\\
ECE \& ASRI, Seoul National University, Korea \hspace{1.0cm}
\\
{\small \texttt {\{redarknight,mks0601,jkpark0825,kyoungmu\}@snu.ac.kr}}
}


\maketitle

\begin{abstract}
We consider the problem of recovering a single person's 3D human mesh from in-the-wild crowded scenes.
While much progress has been in 3D human mesh estimation, existing methods struggle when test input has crowded scenes.
The first reason for the failure is a domain gap between training and testing data.
A motion capture dataset, which provides accurate 3D labels for training, lacks crowd data and impedes a network from learning crowded scene-robust image features of a target person.
The second reason is a feature processing that spatially averages the feature map of a localized bounding box containing multiple people.
Averaging the whole feature map makes a target person's feature indistinguishable from others.
We present 3DCrowdNet that firstly explicitly targets in-the-wild crowded scenes and estimates a robust 3D human mesh by addressing the above issues.
First, we leverage 2D human pose estimation that does not require a motion capture dataset with 3D labels for training and does not suffer from the domain gap.
Second, we propose a joint-based regressor that distinguishes a target person's feature from others.
Our joint-based regressor preserves the spatial activation of a target by sampling features from the target's joint locations and regresses human model parameters.
As a result, 3DCrowdNet learns target-focused features and effectively excludes the irrelevant features of nearby persons.
We conduct experiments on various benchmarks and prove the robustness of 3DCrowdNet to the in-the-wild crowded scenes both quantitatively and qualitatively.
Codes are available here: \url{https://github.com/hongsukchoi/3DCrowdNet_RELEASE}
\end{abstract}
\section{Introduction}

Extensive research has been committed to reconstructing an accurate 3D human mesh, which represent both the pose and shape of a human, from a single image.
However, 3D human mesh estimation from in-the-wild crowded scenes has been barely studied, despite their common presence.
Consequently, most of the previous works show results on scenes without inter-person occlusion and provide inaccurate results on crowded scenes. 
The inter-person occlusion is the essential challenge of in-the-wild crowded scenes, and many practical applications including abnormal behavior detection~\cite{gatt2019detecting} and person re-identification~\cite{narayan2018re} encounter such situations.
This paper investigates the limitation of the current literature and proposes a novel method for robust 3D human mesh estimation from in-the-wild crowded scenes.

{\centering
\begin{table}[t]
\begin{tabular}{c}

\begin{subfigure}{0.45\textwidth}\centering\includegraphics[width=\columnwidth]{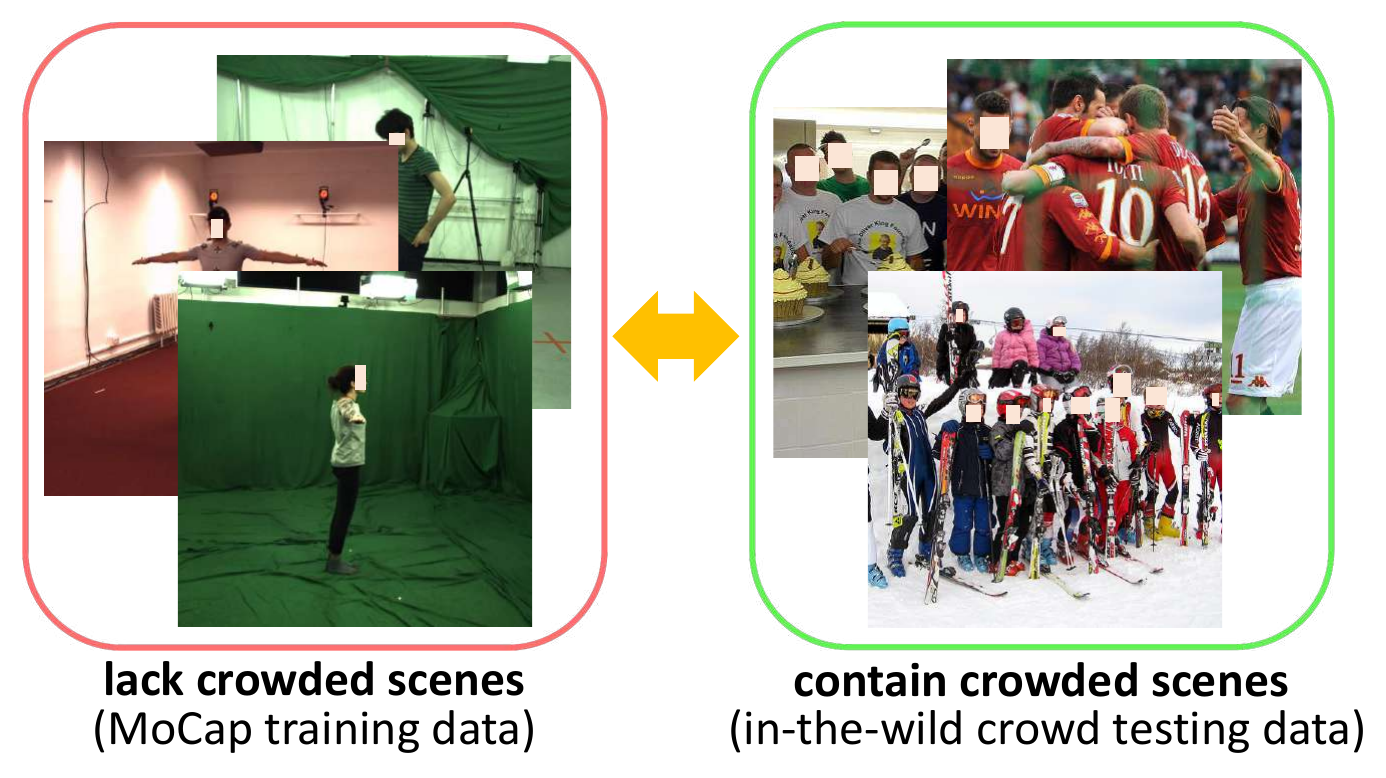}\caption{Domain gap (occlusion/pose/appearance, etc)}\label{fig:domain_gap}\vspace{0.2em}\end{subfigure} \\ 
\begin{subfigure}{0.45\textwidth}\centering\includegraphics[width=\columnwidth]{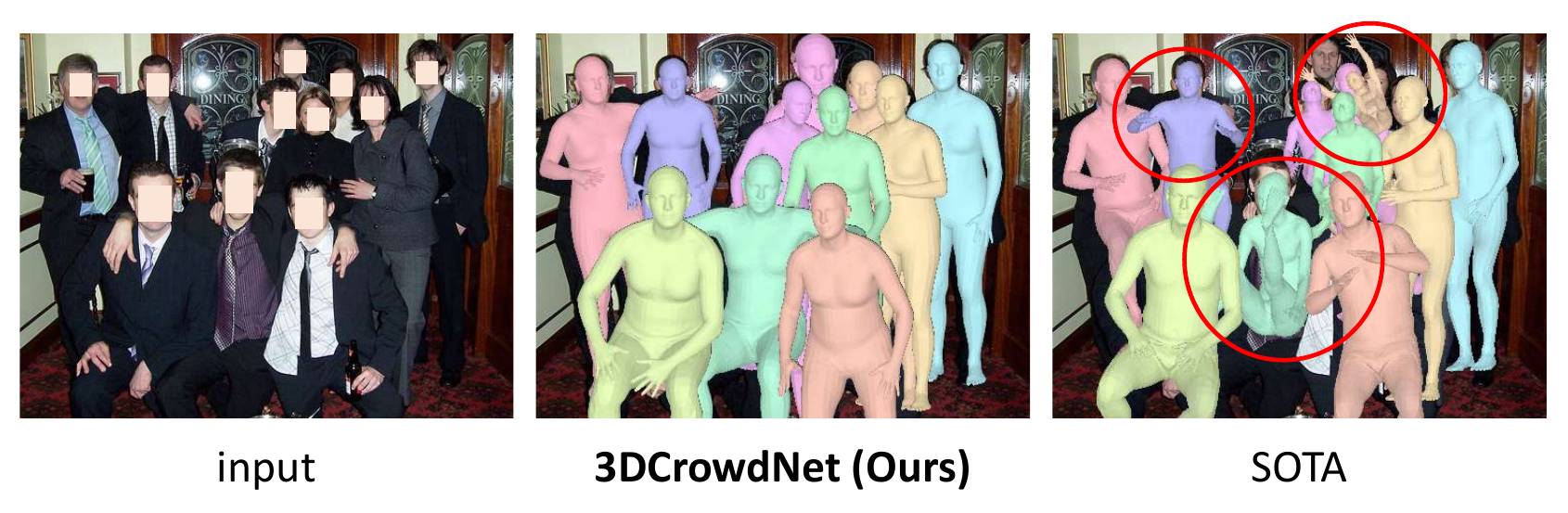}\caption{Qualitative comparison}\label{fig:sota_comparison}\end{subfigure}

\end{tabular}

\captionof{figure}
{
3DCrowdNet resolves (a) a domain gap issue in estimating a 3D human mesh from in-the-wild crowded scenes. Due to the large domain gap between motion capture data and in-the-wild crowd data, (b) existing state-of-the-art methods such as SPIN~\cite{kolotouros2019learning} produce inaccurate results, while 3DCrowdNet gives an accurate 3D human mesh despite severe inter-person occlusion. We conceal a person's face in this paper to abide by the ethical policy.
}
\label{fig:front_figure}
\end{table}
}
\setcounter{table}{0}

The currently dominant training strategy for human mesh recovery is mixed-batch training.
It composes a mini-batch with one-half data from a motion capture (MoCap) 3D dataset~\cite{ionescu2014human3,mehta2017monocular} and the other from an in-the-wild 2D dataset~\cite{lin2014mscoco}.
To use the 2D dataset for supervision, 3D joints regressed from a predicted mesh are projected onto the image plane, and the distance with 2D annotations is computed.
This way of mixing 3D and 2D data is well known to improve accuracy and generalization~\cite{kanazawa2018end,kolotouros2019learning} by implicitly inducing a neural network to benefit from accurate 3D annotations of the 3D data and diverse image appearances in the 2D data.
The dominant approach of recent works~\cite{kolotouros2019learning,georgakis2020hierarchical,choi2021beyond} is a model-based approach using a global feature vector, which obtains the feature vector with a deep convolutional neural network (CNN) and regresses the human model parameters (\textit{e.g.} SMPL~\cite{loper2015smpl}) from it.
First, they crop an image using a bounding box of a target person detected from off-the-shelf human detectors~\cite{he2017mask}.
Then they process the target's cropped image with a deep CNN and perform a global average pooling to obtain the global feature vector.
The global feature vector is fed to a Multi-Layer Perceptron (MLP)-based regressor that regresses the mesh parameters.
The 3D meshes are obtained by forwarding the parameters to the human model layers.

While the recent works have shown reasonable results on standard benchmarks~\cite{ionescu2014human3,von20183dpw} based on the two wheels of the current literature, in-the-wild crowded scenes remain insurmountable due to the following two reasons. 
First, a large domain gap between training data from MoCap datasets and testing data from in-the-wild crowded scenes hinders a deep CNN from extracting proper image features of a target person. 
The domain gap arises from the presence of a human crowd, which entails diverse inter-person occlusion, interacting body poses, and indistinguishable cloth appearances (Figure~\ref{fig:domain_gap}).
The mixed-batch training alone is insufficient to overcome the domain gap, and existing methods struggle to acquire robust image features from in-the-wild crowded scenes, and produce inaccurate meshes (Figure~\ref{fig:sota_comparison}).
Intuitively, this tells us that external guidance robust to the domain gap is required for a crowded scene-robust image feature, in addition to the mixed-batch training.



Next, the global average pooling on a deep CNN feature collapses the spatial information that distinguishes a target person's feature from others.
In-the-wild crowded scenes often involve overlapping people and inaccurate human bounding boxes.
Thus, a bounding box of a target inevitably includes non-target people.
A deep CNN feature retains features of these non-target people, and the global average pooling makes a target person’s feature indistinguishable from others.
This confuses a regressor and makes it difficult to capture an accurate 3D pose of a target person.
For instance, the regressor may miss human parts occluded by another person or predict a different person's pose.

In this regard, we present 3DCrowdNet, a novel network that learns to estimate a single person's robust 3D human mesh from in-the-wild crowded scenes.
This study is one of the earliest works that explicitly tackle 3D human mesh estimation of a target person in a crowd.
3DCrowdNet addresses the two issues of previous works in two folds.
First, we resolve the domain gap by explicitly guiding a deep CNN to extract a crowded scene-robust image feature using an off-the-shelf 2D pose estimator.
Unlike methods targeting 3D geometry, the 2D pose estimator does not require depth supervision and is not trained on a MoCap dataset.
Instead, it is trained only on in-the-wild datasets~\cite{li2019crowdpose,qiu2020peeking} that have images containing human crowds and suffers less from a domain gap regarding the inference on crowded scenes.
Consequently, the 2D pose estimator's outputs provide strong evidence of a target person and help 3DCrowdNet pay attention to a target's feature despite the challenges in in-the-wild crowded scenes.

\begin{figure*}
\begin{center}
\includegraphics[width=0.9\linewidth]{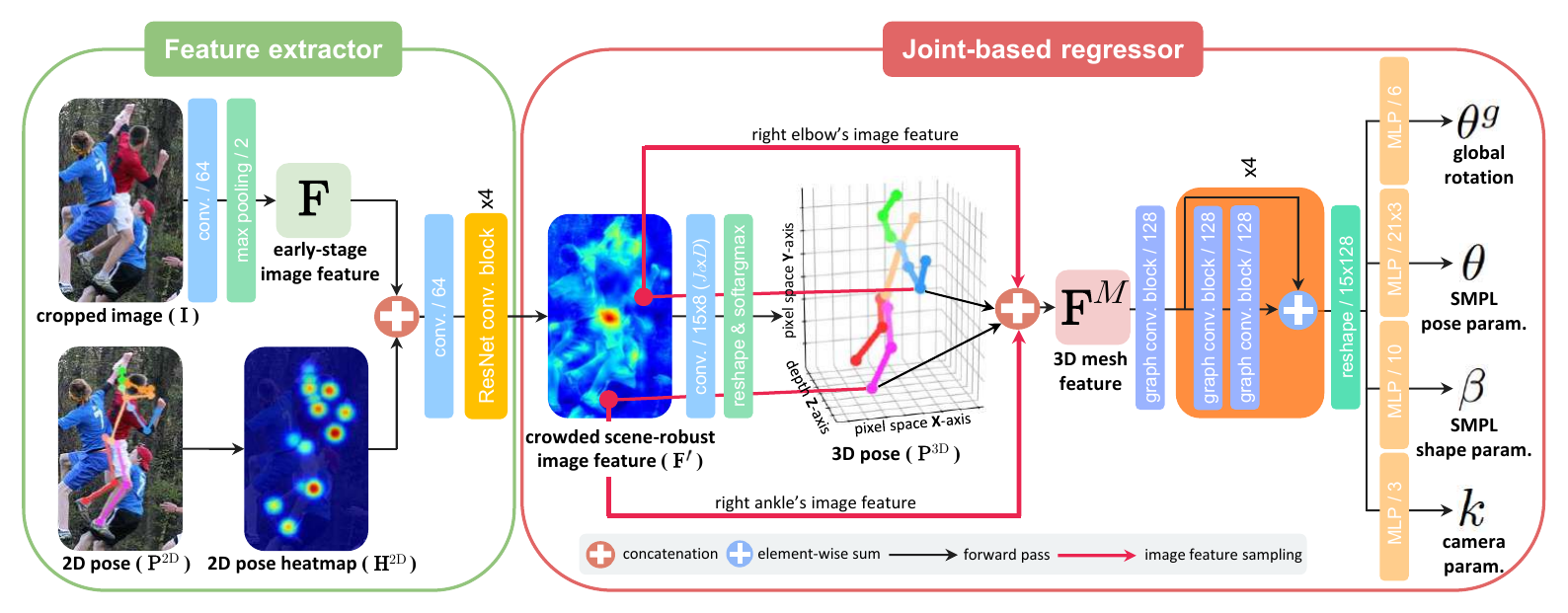}
\end{center}
   \caption{
    Overview of 3DCrowdNet. It resolves the domain gap by explicitly guiding a deep CNN to extract a crowded scene-robust feature using an off-the-shelf 2D pose estimator. Then, it distinguishes a target person from others by preserving the target's spatial activation with a joint-based regressor and regresses SMPL~\cite{loper2015smpl} parameters. The parameters are fed to the SMPL layer to get a 3D mesh. For simplicity, we show image feature sampling on only two joints. The numbers in network layers indicate the output channel dimension. The number in the max pooling layer indicates a stride size. The graph convolutional blocks’ channel dimension is defined per joint.
   }
\label{fig:architecture}
\end{figure*}

Second, we propose a joint-based regressor that does not blow away the spatial activation of a target person in a feature map with the global average pooling.
The joint-based regressor first predicts the spatial locations of joints. 
Then, it samples image features from a deep CNN feature map with the locations.
In particular, we keep the sampling area small to exclude features of non-target people.
The target person's feature is distinguished from others, and human model parameters are regressed from the sampled image features.
The joint-based regressor differs from the previous regressors that evenly aggregate people's features regardless of the target.
Figure~\ref{fig:architecture} depicts the overview of 3DCrowdNet.

Note that 3DCrowdNet substantially differs from prior works~\cite{martinez2017simple,choi2020p2m} that directly lift 2D estimation outputs to 3D- (a) we focus on producing and leveraging image features of a target person in human crowds, and
(b) such image features help 3DCrowdNet to resolve the depth and shape ambiguity of a target person, from which the 2D estimation outputs inherently suffer.
Thus, we argue that this work takes a step towards accurate 3D human mesh estimation from in-the-wild crowded scenes by distinguishing image features of a target person in densely interacting crowds, which is highly challenging but important.
The experiments show that 3DCrowdNet significantly outperforms the previous 3D human mesh estimation methods on in-the-wild crowded scenes.
Also, it achieves state-of-the-art accuracy in multiple 3D benchmarks~\cite{mehta2018single,joo2017panoptic,von20183dpw}.
Extensive qualitative results are presented in the main manuscript and supplementary material.
Our contributions can be summarized as follows:
\begin{itemize}
\item We present 3DCrowdNet, the first approach to 3D human mesh recovery from in-the-wild crowded scenes. 
It effectively processes image features of a target person in a crowd, which is essential for accurate 3D pose and shape reconstruction. 
\item It extracts crowded scene-robust image features by resolving the domain gap with a 2D pose estimator.
\item It distinguishes a target person's image features from others using a joint-based regressor.
\item 3DCrowdNet significantly outperforms previous methods on in-the-wild crowded scenes both quantitatively and qualitatively, and achieves state-of-the-art 3D pose and shape accuracy on multiple 3D benchmarks.
\end{itemize}

\section{Related works}

\noindent\textbf{2D human pose estimation from crowded scenes.}
Early works of 2D human pose estimation did not explicitly target crowded scenes. However, their methods are related to diverse challenges of in-the-wild crowded scenes, such as overlapping human bounding boxes, human detection error, and inter-person occlusion.
There are two major approaches, namely bottom-up and top-down approaches.
Bottom-up methods~\cite{pishchulin2016deepcut,cao2017realtime,newell2016associative} first detect all joints of the people, and group them to each person.
Top-down methods~\cite{he2017mask,papandreou2017towards,chen2018cascaded} first detect all human bounding boxes, and apply a single-person 2D pose estimation method to each person.
Top-down methods generally achieve higher accuracy on traditional 2D pose benchmarks such as MSCOCO~\cite{lin2014mscoco}, but underperform on crowded scene benchmarks~\cite{li2019crowdpose,zhang2019pose2seg} than bottom-up methods due to the human detection issues. 

Recently, a few works explicitly addressed crowded scenes 2D pose estimation and reported good accuracy on crowded scene benchmarks.
~\cite{li2019crowdpose} combined top-down and bottom-up approaches using joint-candidate single person pose estimation and global maximum joints association.
~\cite{cheng2020higherhrnet} proposed to learn scale-aware representations using high-resolution feature pyramids.
~\cite{jin2020differentiable} made a grouping process of the bottom-up approach differentiable using a graph neural network. 
~\cite{qiu2020peeking} refined invisible joints' prediction using an image-guided progressive graph convolutional network.

\noindent\textbf{3D human geometry estimation from crowded scenes.}
Several methods~\cite{moon2019camera,wang2020hmor,zhen2020smap} have shown reasonable results on multi-person 3D benchmarks~\cite{joo2017panoptic,mehta2017monocular}.
However, their focus was on absolute depth estimation of each person, and few works have addressed the inter-person occlusion to estimate robust 3D geometry, such as 3D human pose (\textit{i.e.} 3D joint coordinates) and meshes, from in-the-wild crowded scenes.
XNect~\cite{mehta2020xnect} proposed an occlusion-robust method that can be applied to crowded scenes.
However, it did not focus on resolving the domain gap.
It integrated 2D/3D branches into a single system and trained it on a MoCap dataset~\cite{mehta2018single}, which barely contains inter-person occlusions.
Also, it requires a particular joint (\textit{i.e.} neck) must be visible for human detection.
On the contrary, our key idea is leveraging~\textit{external} 2D pose estimators that are not trained on MoCap data, to alleviate the domain gap between MoCap training data and in-the-wild crowd testing data.
In addition, 3DCrowdNet reconstructs full 3D human pose and shape from diverse partially invisible people in crowded scenes.

ROMP\cite{sun2021monocular} introduced a bottom-up method for multi-person 3D mesh recovery that can be applied to crowded scenes.
It estimates a body center heatmap and a mesh parameter map, and samples each person's mesh parameters from the parameter map using center locations regressed from the heatmap.
While the method provides better results on crowded scenes than previous methods, it could still suffer from the domain gap between MoCap training data and testing data from in-the-wild crowded scenes.
Also, solely relying on the body center estimation to distinguish a target from others could be unstable in cases of occlusion on the body center.
On the other hand, 3DCrowdNet explicitly tackles the domain gap issue with crowded-scene robust 2D poses.
Also, we utilize cues from multiple 2D joint locations of the target and refine image features sampled from the locations to handle diverse inter-person occlusion, including occlusion on the body center.


\noindent\textbf{2D geometry to 3D human mesh estimation.}
~\cite{choi2020p2m,zhang2020learning,song2020human,sengupta2020synthetic} proposed methods that only take 2D geometry without images, such as 2D joint locations, for SMPL parameter regression.
While the methods can benefit from 2D estimators robust to in-the-wild crowded scenes, they have two limitations.
First, they cannot correct inaccurate 2D input compared to the actual person in images.
Instead, they produce the most plausible outputs for the given 2D input, not the 3D pose and shape that best describes the person in images.
Second, they do not benefit from image features with rich depth and 3D shape cues of a target person.
The cues include subtle light reflection and shadows.
2D geometry hardly contains such cues and could lead to inaccurate 3D human mesh estimation. 
On the contrary, 3DCrowdNet reconstructs accurate 3D human meshes from possibly inaccurate 2D poses utilizing image features.
Also, we focus on extracting the crowded scene-robust image feature of a target person using the 2D pose, rather than directly lifting 2D to 3D as the prior works.


\section{3DCrowdNet}

\subsection{3DCrowdNet architecture}
As shown in Figure~\ref{fig:architecture}, our architecture comprises a feature extractor followed by a joint-based regressor.
The feature extractor is based on ResNet-50~\cite{he2016deep}, and the joint-based regressor is based on~\cite{moon2020pose2pose,liu2020comprehensive}.
Our network's output is SMPL~\cite{loper2015smpl} parameters, and a single person's 3D mesh is obtained by feeding the parameters to the SMPL layer.

\noindent\textbf{Feature extractor.}
\label{section:prepare_2dpose}
The feature extractor takes a 2D pose and an image as input.
The 2D pose is 2D joint coordinates $\mathbf{P}^{\text{2D}} \in \mathbb{R}^{J \times 2}$ predicted by bottom-up off-the-shelf 2D pose estimators~\cite{cao2017realtime,cheng2020higherhrnet}.
$J$ denotes the number of human joints, and it can vary among different 2D pose estimators.
During training, we add realistic errors on the ground truth (GT) 2D pose following~\cite{moon2019posefix,choi2020p2m} to mimic erroneous 2D pose outputs in test time, and the noisy 2D pose is used as our input $\mathbf{P}^{\text{2D}}$.
We provide the 2D pose $\mathbf{P}^{\text{2D}}$ as a heatmap representation $\mathbf{H}^{\text{2D}} \in \mathbb{R}^{J_s \times 64 \times 64}$ to the feature extractor by making a Gaussian blob on the 2D joint coordinates.
$J_s=30$ indicates the number of joints in a superset of joint sets defined by multiple datasets.
We assign don't-care values to the undefined joints and joint predictions with low confidence in inference time, by multiplying zero to the corresponding joint's heatmap.
Modeling don't-care values based on the superset of joints and heatmaps enables 3DCrowdNet to perform inference from various human joint sets with a single network and handle diverse input such as 2D poses with missing joints due to truncation and occlusion.

The feature extractor uses the 2D pose heatmap $\mathbf{H}^{\text{2D}}$ of a target person as guidance and pays attention to the spatial region of a target in a crowd.
First, it obtains an early-stage image feature of ResNet $\mathbf{F} \in \mathbb{R}^{C \times 64 \times 64}$ from a cropped image $\mathbf{I} \in \mathbb{R}^{3 \times 256 \times 256}$.
$C=64$ is the channel dimension, and $\mathbf{I}$ is acquired by cropping and resizing a bounding box area, derived from the 2D pose $\mathbf{P}^{\text{2D}}$.
Second, it concatenates $\mathbf{F}$ and $\mathbf{H}^{\text{2D}}$ along the channel dimension.
The concatenated feature is processed by a 3-by-3 convolution block, which keeps the feature's height and width but changes the channel dimension to $C$. 
Finally, the feature with $C$ channels is fed back to the remaining part of ResNet, where the output is 
a crowded scene-robust image feature $\mathbf{F^\prime} \in \mathbb{R}^{C^\prime \times 8 \times 8}$.
$C^\prime=2048$ is the channel dimension. 

\noindent\textbf{Joint-based regressor.} 
The joint-based regressor first recovers 3D joint coordinates $\mathbf{P}^{\text{3D}} \in \mathbb{R}^{J_c \times 3}$ from $\mathbf{F^\prime}$.
$J_c=15$ denotes the number of joints in the intersection of joint sets defined by multiple datasets.
($x$,$y$) values of $\mathbf{P}^{\text{3D}}$ are defined in a 2D pixel space, and $z$ value of $\mathbf{P}^{\text{3D}}$ represents root joint-relative depth.
A 1-by-1 convolutional layer outputs a 3D heatmap $\mathbf{H}^{\text{3D}} \in \mathbb{R}^{J_c \times D \times 8 \times 8}$ from $\mathbf{F^\prime}$ after predicting a $J_cD$ dimensional 2D feature map and reshaping it to the 3D heatmap.
$D=8$ decides a descritizated size of depth.
$\mathbf{P}^{\text{3D}}$ is computed from $\mathbf{H}^{\text{3D}}$, using soft-argmax operation~\cite{sun2017compositional}.
As the soft-argmax computes continuous coordinates from a discretized grid, we observed that a heatmap with a low resolution like $\mathbf{H}^{\text{3D}}$ gives similar accuracy compared to upsampled ones, while requiring less computational costs.

Next, the joint-based regressor estimates global rotation of a person $\theta^g \in \mathbb{R}^{3}$, SMPL body rotation parameters $\theta \in \mathbb{R}^{21 \times 3}$, SMPL shape parameters $\beta \in \mathbb{R}^{10}$, and camera parameters $k \in \mathbb{R}^{3}$ for projection.
First, image features per joint are sampled from $\mathbf{F^\prime}$ using the ($x$,$y$) pixel positions of $\mathbf{P}^{\text{3D}}$.
We use bilinear interpolation, since the ($x$,$y$) pixel positions are not in discretized values.
The prediction confidence of $\mathbf{P}^{\text{3D}}$ is sampled from $\mathbf{H}^{\text{3D}}$ in the same manner. 
Second, we concatenate the sampled image features, $\mathbf{P}^{\text{3D}}$, and the prediction confidence of $\mathbf{P}^{\text{3D}}$, to attain $\mathbf{F}^{M} \in \mathbb{R}^{J_c \times (C^\prime+3+1)}$.
Last, we process $\mathbf{F}^{M}$ using a graph convolutional network (GCN), and predict $\theta^g$, VPoser~\cite{pavlakos2019expressive} latent code $z$, $\beta$, and $k$ from output features of GCN with separate MLP layers.
$\theta$ is decoded from $z$.
The GCN shows faster convergence during training than an MLP network, and we think the reason lies behind the character of $\theta$.
$\theta$ is parent joint-relative joint rotations, and the GCN can exploit the human kinematic prior different from an MLP.
For example, the GCN can implicitly learn the valid range of each parent joint-relative joint leveraging the relationship between human joints.

For the graph convolutional network, we use the joint-specific graph convolution~\cite{liu2020comprehensive} that learns separate weights for each graph vertex.
We define learnable weight matrices $\{W_j \in \mathbb{R}^{C_{\text {out}} \times C_{\text {in}}}\}_{j=1}^{J_c}$ for all joints of each graph convolution layer, where $C_{\text {in}}$ and $C_{\text {out}}$ denotes input and output channel dimensions, respectively.
Then, the output graph feature of joint $j$ is derived as $\mathbf{F}_j^{\text {out}} = \sigma_\text{ReLU}(\sum_{i \in \hat{\mathcal{N}}_j} \tilde{a}_{ji} \sigma_\text{BN}( W_i \mathbf{F}_i^{\text {in}}))$, where $\mathbf{F}_i^{\text {in}}$ is the input graph feature of joint $i$.
$\sigma_\text{ReLU}$ and $\sigma_\text{BN}$ denotes ReLU activation function and 1D batch normalization~\cite{ioffe2015batch}, respectively.
$\hat{\mathcal{N}}_j$ is defined as $\mathcal{N}_j \cup \{j\}$, where $\mathcal{N}_j$ denotes neighbors of a vertex $j$.
$\tilde{a}_{ji}$ is an entry of the normalized adjacency matrix $\tilde {\mathbf{A}}$ at $(j,i)$, where $\tilde {\mathbf{A}}=\mathbf{D}^{-\frac{1}{2}}(\mathbf{A}+\mathbf{I}) \mathbf{D}^{-\frac{1}{2}}$.
$\mathbf{A} \in \{0,1\}^{J_c \times J_c}$ is the adjacency matrix constructed based on the human skeleton hierarchy and fixed during the training and testing stages.
The definition of the human skeleton hierarchy is depicted in the supplementary material.

\subsection{Network training}
The feature extractor and joint-based regressor are integrated and trained end-to-end.
We use both pseudo-GT SMPL fits obtained by fitting frameworks~\cite{pavlakos2019expressive,moon2020neuralannot} and GT annotations from training datasets for supervision following~\cite{kolotouros2019learning}.
Our overall objective is defined as follows:

\begin{equation}
L = L_{\text{pose}} + L_{\text{mesh}},
\end{equation}
where $L_{\text{pose}}$ computes the L1 distance between the predicted $\mathbf{P}^{\text{3D}}$ and the (pseudo) GT, and $L_{\text{mesh}}$ denotes the loss function for predicted SMPL parameters.
$L_{\text{mesh}}$ is defined as

\begin{equation}
L_{\text{mesh}} = L_{\text{param}} + L_{\text{pose}^\prime},
\end{equation}
where $L_{\text{param}}$ computes the L1 distance between the predicted $\theta^g$, $\theta$, and $\beta$, and the pseudo-GT parameters; 
$L_{\text{pose}^\prime}$ indicates the L1 distance loss of joints regressed from predicted meshes.
To supervise with 2D annotations~\cite{lin2014mscoco,andriluka2014mpii}, predicted joints are projected by camera parameters $k$. 

\subsection{Implementation detail}
PyTorch~\cite{paszke2017automatic} is used for implementation. 
We initialize the weights of ResNet~\cite{he2016deep} with the pre-trained weights from~\cite{xiao2018simple}.
It shows faster convergence during training.
We use Adam optimizer~\cite{kingma2014adam} with a mini-batch size of 64.
The initial learning rate is $10^{-4}$.
The model is trained for $6$ epochs, and the learning rate is reduced by a factor of 10 after the $3$th and $5$th epochs.
We use four NVIDIA RTX 2080 Ti GPUs for training, and it takes about 9 hours on average.
We will release the codes for more details.
\section{Experiment}
\label{section:experiment}
\subsection{Datasets}

\noindent\textbf{Training sets.}
We use Human3.6M~\cite{ionescu2014human3}, MuCo-3DHP~\cite{mehta2018single}, MSCOCO~\cite{lin2014mscoco}, MPII~\cite{andriluka2014mpii}, and CrowdPose~\cite{li2019crowdpose} for training.
Only the training sets of the datasets are used, following the standard split protocols.

\noindent\textbf{Testing sets.}
We report accuracy on MuPoTS~\cite{mehta2018single}, CMU-Panoptic~\cite{joo2017panoptic}, 3DPW~\cite{von20183dpw}, and 3DPW-Crowd.
MuPoTS is a multi-person test benchmark captured from indoor and outdoor environments, starring 3 to 4 people.
CMU-Panoptic is a large-scale multi-person dataset captured from the Panoptic studio.
Following~\cite{zanfir2018deep,jiang2020coherent}, we pick four sequences presenting 3 to 7 people socializing each other for the evaluation.
3DPW is a widely-used 3D benchmark captured from an in-the-wild environment, and we use the test set of 3DPW following the official split protocol.
3DPW-Crowd is a subset of 3DPW and is used to evaluate the a method's robustness to in-the-wild crowded scenes.
Refer to more details below about its necessity.

\begin{figure}[t]
\begin{center}
\includegraphics[width=0.9\linewidth]{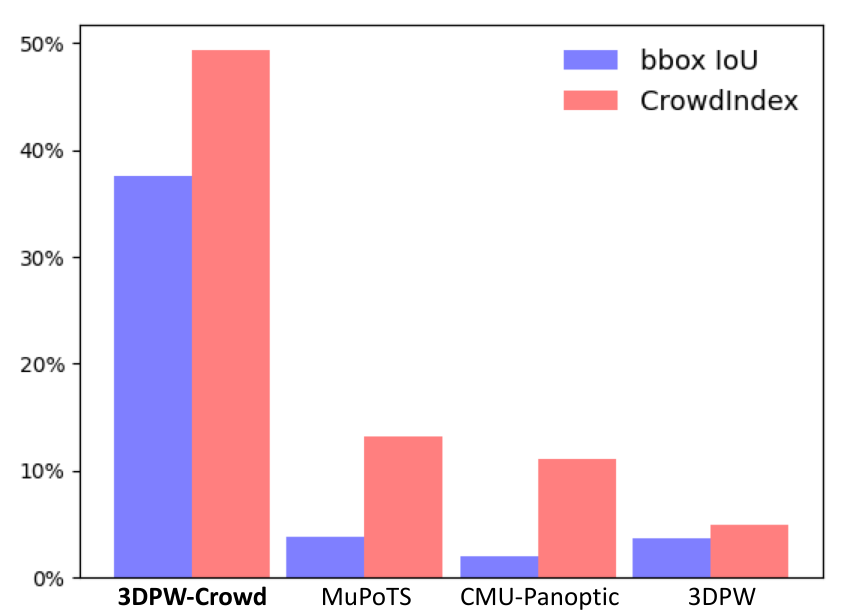}
\vspace*{-1em}
\end{center}
   \caption{
   We curate 3DPW-Crowd, a subset of 3DPW, which has much higher bounding box IoU and CrowdIndex~\cite{li2019crowdpose} than other 3D benchmarks. CrowdIndex measures other people's joints' ratio over each person's joints in a bounding box. 
   }
\label{fig:data_statistics}
\vspace*{-1em}
\end{figure}

\subsection{Evaluation protocols}

\noindent\textbf{Evaluation on crowded scenes: 3DPW-Crowd and CrowdPose.}
As CrowdPose~\cite{li2019crowdpose} addressed, the principal obstacle of pose estimation from crowded scenes is not the number of people, but the inter-person occlusion in a crowd.
Thus, MuPoTS~\cite{mehta2018single} and CMU-Panoptic~\cite{joo2017panoptic} have limitations for the evaluation on in-the-wild crowded scenes, not only because they are not in-the-wild data, but also because they show limited interaction.

To overcome the limitations, we propose 3DPW-Crowd to numerically measure a method's robustness on in-the-wild crowded scenes. 
It contains hugging and dancing sequences that have considerably higher average intersection over union (IoU) of bounding boxes and CrowdIndex~\cite{li2019crowdpose} among 3D benchmarks as shown in Figure~\ref{fig:data_statistics}.
We name the subset as 3DPW-Crowd, since it reveals the challenges of in-the-wild crowded scenes, such as overlapping bounding boxes and severe inter-person occlusion.
More details about 3DPW-Crowd are in the supplementary material.
We also provide extensive qualitative comparison between different methods on the test set of CrowdPose~\cite{li2019crowdpose} in this manuscript and supplementary material.

\noindent\textbf{Evaluation metrics.}
We report 3D pose and 3D shape evaluation metrics.
For the 3D pose evaluation, we use mean per-joint position error (MPJPE), Procrustes-aligned mean per-joint position error (PA-MPJPE), and 3DPCK proposed in~\cite{mehta2017monocular}.
Following SPIN~\cite{kolotouros2019learning}, we use the 3D joint coordinates regressed from a 3D mesh as predictions.
For the 3D shape evaluation, we use mean per-vertex position error (MPVPE).
All errors are measured after aligning root joints of GT and estimated human body meshes.

\subsection{Ablation study}

We carry out the ablation study on 3DPW-Crowd.
We use HigherHRNet~\cite{cheng2020higherhrnet}'s 2D pose outputs in Table~\ref{table:regressor_ablation},~\ref{table:sampling_area_ablation}, and~\ref{table:3dposenet_ablation}.

\begin{figure}[t]
\begin{center}
\includegraphics[width=0.85\linewidth]{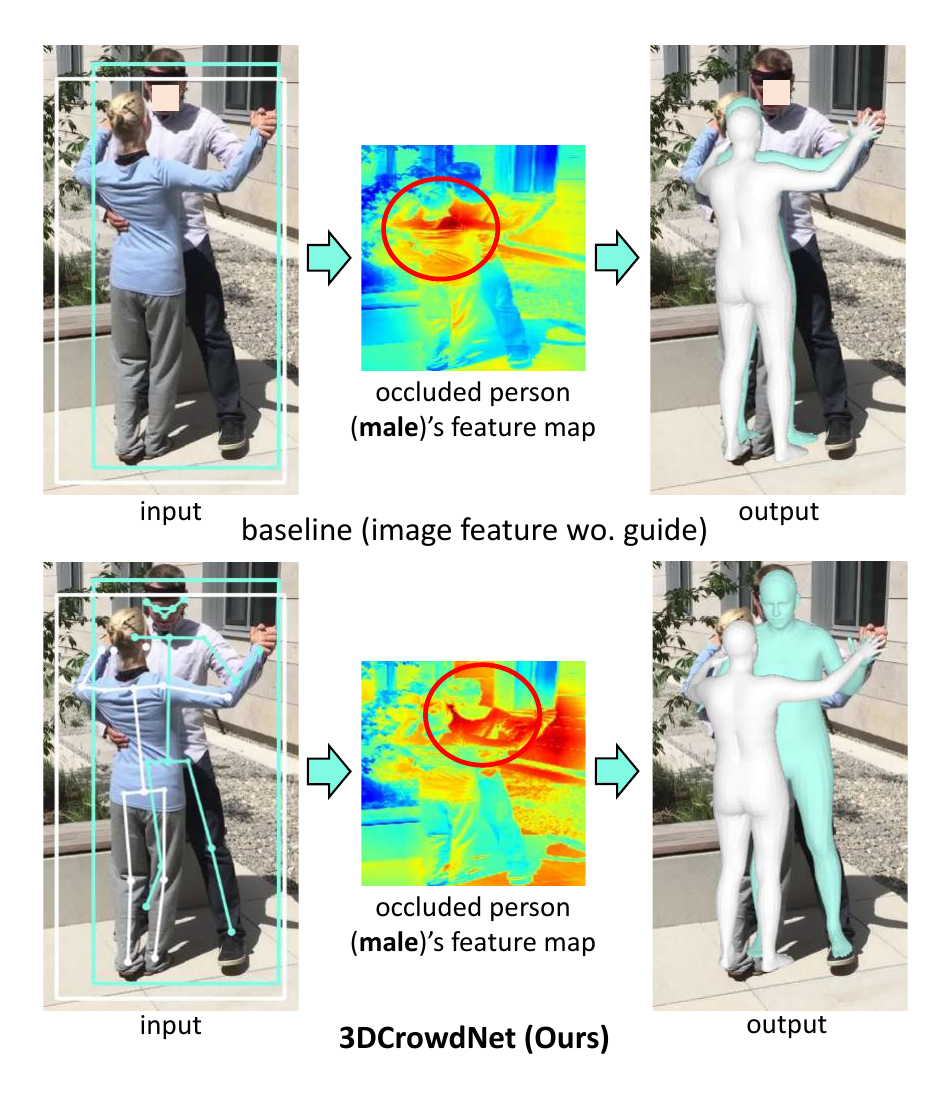}
\end{center}
\vspace*{-1em}
   \caption{
   Comparison between the baseline that takes only image features and 3DCrowdNet. The baseline gives stronger attention to an occluding person (female) instead of an occluded person (male), and produces a wrong 3D mesh. 3DCrowdNet pays attention to the target male and recovers an accurate 3D mesh.
   }
\label{fig:activation_ablation}
\vspace*{-0.5em}
\end{figure}

\begin{table}[t]
\small
\centering
\setlength\tabcolsep{1.0pt}
\def\arraystretch{1.1}
\begin{tabular}{C{5.45cm}|C{1.15cm}C{1.6cm}}
\specialrule{.1em}{.05em}{.05em}
input feature & MPJPE$\downarrow$ & PA-MPJPE$\downarrow$ \\ \hline
image feature wo. guide & 109.6 & 63.3 \\
\cellcolor{Gray}\textbf{crowded scene-robust image feature} & \cellcolor{Gray}\textbf{85.8} & \cellcolor{Gray}\textbf{55.8} \\
\specialrule{.1em}{.05em}{.05em}
\end{tabular}
\vspace*{-1.0em}
    \caption{Ablation on the input image features.}
\label{table:input_ablation_3dpw_crowd}
\vspace*{-0.5em}
\end{table}

\noindent\textbf{Crowded scene-robust image feature.}
Table~\ref{table:input_ablation_3dpw_crowd} shows the effectiveness of the crowded scene-robust image feature.
The baseline network in the first row crops an image using a GT bounding box and extracts image features without any guidance as in previous methods.
The significant error drop in the table proves that the 2D pose can produce crowded scene-robust image features, and the image features are critical to estimate an accurate mesh from the in-the-wild crowded scenes.
We further validate our statement that the 2D pose can produce crowded scene-robust image features in Figure~\ref{fig:activation_ablation}.
3DCrowdNet activates the occluded target male's spatial region, unlike the baseline network, and successfully distinguishes him from the other.
As a result, 3DCrowdNet estimates an accurate mesh of the occluded target male, while the baseline predicts a mesh of the occluding female.
We conclude that for the 3D mesh estimation on in-the-wild crowded scenes, the domain difference of MoCap train data is the bottleneck, and our idea to exploit the robustness of 2D pose estimators, which do not use MoCap train data, is valid.



\begin{table}[t]
\small
\centering
\setlength\tabcolsep{1.0pt}
\def\arraystretch{1.1}
\begin{tabular}{C{4.0cm}|C{1.3cm}C{1.8cm}}
\specialrule{.1em}{.05em}{.05em}
parameter regressor type & MPJPE$\downarrow$ & PA-MPJPE$\downarrow$ \\ \hline

SPIN-style regressor & 89.0 &  59.5 \\
\cellcolor{Gray}\textbf{joint-based regressor {\scriptsize(Ours)}}  &  \cellcolor{Gray}\textbf{85.8} &  \cellcolor{Gray}\textbf{55.8} \\
\specialrule{.1em}{.05em}{.05em}
\end{tabular}
\vspace*{-1.0em}
\caption{Ablation of the regressor types.
}
\label{table:regressor_ablation}
\vspace*{-0.5em}
\end{table}

\begin{table}[t]
\small
\centering
\setlength\tabcolsep{1.0pt}
\def\arraystretch{1.1}
\begin{tabular}{C{4.2cm}|C{1.3cm}C{1.8cm}}
\specialrule{.1em}{.05em}{.05em}
sampling area & MPJPE$\downarrow$ & PA-MPJPE$\downarrow$ \\ \hline
whole feature map & 89.1 & 57.8 \\
\cellcolor{Gray}5-by-5 grid around point & \cellcolor{Gray}88.2 &  \cellcolor{Gray}57.6 \\
\textbf{point {\scriptsize(Ours)}} & \textbf{85.8} & \textbf{55.8} \\
\specialrule{.1em}{.05em}{.05em}
\end{tabular}
\vspace*{-1.0em}
\caption{Ablation on the sampling area of image features.
}
\label{table:sampling_area_ablation}
\vspace*{-1.5em}
\end{table}

\noindent\textbf{Joint-based regressor.}
Table~\ref{table:regressor_ablation} shows that the joint-based regressor outperforms the SPIN~\cite{kolotouros2019learning}-style regressor, the dominant model-based approach in the current literature, on 3DPW-Crowd.
The results prove that preserving the spatial activation of a target person in a deep CNN feature map is essential.
SPIN-style regressor shows lower accuracy, since it makes a target person's feature indistinguishable from others by collapsing the spatial information with a global average pooling. 
We further validate our argument in Table~\ref{table:sampling_area_ablation}.
Originally, our joint-based regressor samples deep image features from ($x$,$y$) positions of the predicted 3D pose.
When we enlarge the sampling area, the errors increase.
Especially, when the joint-based regressor uses features sampled from the whole feature map, which are the same feature of the SPIN-style regressor, MPJPE becomes similar to that of the SPIN-style regressor.
It indicates that most of the accuracy gain in Table~\ref{table:regressor_ablation} is not from a better network architecture, such as GCN, but from the preservation of the target person's spatial activation.
Keeping an appropriate sampling area to less involve non-target people's image features is important to estimate a robust human mesh from in-the-wild crowded scenes.

We also verify the effectiveness of estimating a 3D pose instead of a 2D pose in Table~\ref{table:3dposenet_ablation}.
The clear accuracy improvement proves that the depth information can be reliably estimated from a 2D pose and image features, and it is beneficial for the accuracy of the final mesh estimation.

\begin{table}
\small
\centering
\setlength\tabcolsep{1.0pt}
\def\arraystretch{1.1}
\begin{tabular}{C{4.2cm}|C{1.3cm}C{1.8cm}}
\specialrule{.1em}{.05em}{.05em}
estimation target & MPJPE$\downarrow$ &  PA-MPJPE$\downarrow$ \\ \hline
\cellcolor{Gray}2D pose & \cellcolor{Gray}88.3 &  \cellcolor{Gray}56.4 \\
\textbf{3D pose (Ours)} & \textbf{85.8} & \textbf{55.8} \\
\specialrule{.1em}{.05em}{.05em}
\end{tabular}
\vspace*{-0.6em}
\caption{Ablation on the intermediate estimation target of the joint-based regressor during training and testing.
}
\label{table:3dposenet_ablation}
\vspace*{-0.5em}
\end{table}

\subsection{Comparison with state-of-the-art methods}
Unless indicated, our 3DCrowdNet is not trained on a CrowdPose~\cite{li2019crowdpose} train set in Table~\ref{table:sota_3dpw_crowd},~\ref{table:sota_mupots}, and~\ref{table:sota_cmu}.
Also, we use less or similar training data than other methods, and the details are in the supplementary material.

\begin{table}[t]
\small
\centering
\setlength\tabcolsep{1.0pt}
\def\arraystretch{1.1}
\begin{tabular}{C{3.4cm}|C{1.3cm}C{2.0cm}C{1.3cm}}
\specialrule{.1em}{.05em}{.05em}
method & MPJPE$\downarrow$ & PA-MPJPE$\downarrow$ & MPVPE$\downarrow$ \\ \hline
SPIN~\cite{kolotouros2019learning} & 121.2 & 69.9 & 144.1\\
\cellcolor{Gray}Pose2Mesh~\cite{choi2020p2m} & \cellcolor{Gray}124.8 & \cellcolor{Gray}79.8 & \cellcolor{Gray}149.5 \\
I2L-MeshNet~\cite{moon2020i2l} & 115.7 & 73.5 & 162.0 \\
\cellcolor{Gray}ROMP~\cite{sun2021monocular}$*$ & \cellcolor{Gray}104.8 & \cellcolor{Gray}63.9 & \cellcolor{Gray}127.8 \\
3DCrowdNet {\scriptsize(Ours)} & 86.8 & 56.1 & 109.7 \\
\cellcolor{Gray}\textbf{3DCrowdNet {\scriptsize(Ours)}$*$} & \cellcolor{Gray}\textbf{85.8} & \cellcolor{Gray}\textbf{55.8} & \cellcolor{Gray}\textbf{108.5} \\
\specialrule{.1em}{.05em}{.05em}
\end{tabular}
\vspace*{-0.6em}
    \caption{Comparison on 3DPW-Crowd between 3DCrowdNet and previous methods. We evaluate other methods with their codes and pre-trained models. $*$ means using CrowdPose~\cite{li2019crowdpose}  for training.}
\label{table:sota_3dpw_crowd}
\vspace*{-1.2em}
\end{table}

\noindent\textbf{3DPW-Crowd.}
We compare our 3DCrowdNet with~\cite{kolotouros2019learning,choi2020p2m,moon2020i2l,sun2021monocular} in Table~\ref{table:sota_3dpw_crowd}.
They are recent state-of-the-art 3D human mesh estimation methods on 3DPW, and publicly released the codes for evaluation.
We make several observations.
First, our approach outperforms SPIN~\cite{kolotouros2019learning}, which takes only the image feature as input and performs a global average pooling on the deep CNN feature map.
The result is coherent with the results in Table~\ref{table:input_ablation_3dpw_crowd} and~\ref{table:regressor_ablation} of our ablation studies.
Next, 3DCrowdNet outperforms ROMP~\cite{sun2021monocular}, a bottom-up method for multi-person 3D mesh estimation.
While ROMP achieves higher accuracy than other methods, we think it still suffers from the domain gap issue.
For example, it needs to learn how to distinguish body centers of people under diverse inter-person occlusion, but MoCap datasets they used rarely contain such data.
On the other hand, 3DCrowdNet explicitly resolves the domain gap using 2D pose input and produces accurate 3D meshes.

\begin{figure*}[t]
\begin{center}
\includegraphics[width=0.87\linewidth]{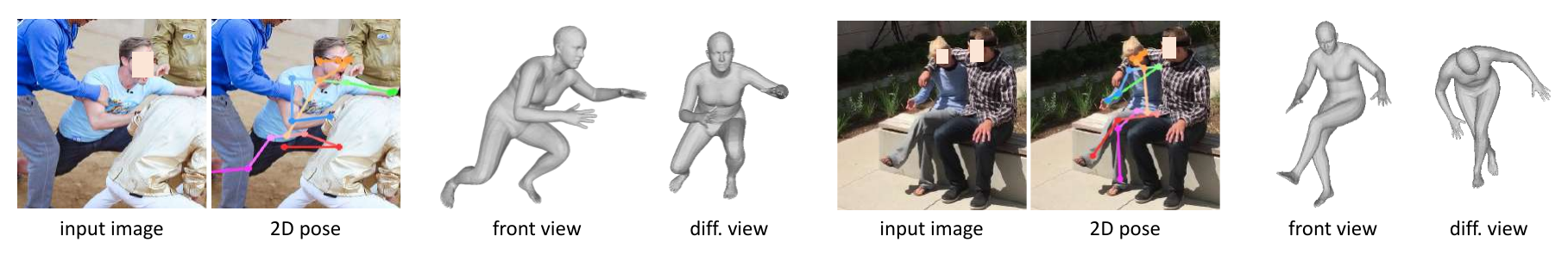}
\end{center}
\vspace*{-2em}
   \caption{
   Visualization from different viewpoints. 3DCrowdNet uses both 2D pose and image features and provide robust results. 
   %
   }
\label{fig:3d_quality}
\vspace*{-1.2em}
\end{figure*}

Last, 3DCrowdNet defeats Pose2Mesh~\cite{choi2020p2m}, a method that can also benefit from crowded-scene robust 2D poses.
We used the same 2D pose predictions of~\cite{cheng2020higherhrnet} for Pose2Mesh and 3DCrowdNet.
The result validates 3DCrowdNet's two strengths over Pose2Mesh.
First, 3DCrowdNet recovers a 3D mesh that best describes a target person, using rich depth and shape cues in images.
On the contrary, Pose2Mesh produces the most plausible 3D mesh for a given 2D pose, and the accuracy depends on it.
Figure~\ref{fig:3d_quality} shows that 3DCrowdNet recovers accurate 3D meshes, even when a 2D pose is inaccurate.
Second, 3DCrowdNet can handle missing joints of 2D pose predictions due to occlusion and truncation owing to don't-care modeling based on the 2D pose's heatmap introduced in Section~\ref{section:prepare_2dpose}.
Pose2Mesh takes the 2D pose as coordinates and cannot cope with the missing joints, common in in-the-wild crowded scenes.
Please also refer to the qualitative comparison in the supplementary material.

\begin{table}[t]
\small
\centering
\setlength\tabcolsep{1.0pt}
\def\arraystretch{1.1}
\begin{tabular}{C{4.8cm}|C{1.3cm}C{1.5cm}}
\specialrule{.1em}{.05em}{.05em}
\multirow{ 2}{*}{method} & \multicolumn{2}{c}{3DPCK$\uparrow$} \\
& All & Matched \\ \hline
SMPLify-X~\cite{pavlakos2019expressive} / {\scriptsize OpenPose}~\cite{cao2017realtime} & 62.8 & 68.0 \\
\cellcolor{Gray}HMR~\cite{kanazawa2018end} / {\scriptsize OpenPose}~\cite{cao2017realtime} & \cellcolor{Gray}66.0 & \cellcolor{Gray}70.9 \\
HMR~\cite{kanazawa2018end} / {\scriptsize Mask R-CNN}~\cite{he2017mask} & 65.6 & 68.6 \\
\cellcolor{Gray}Jiang~\etal~\cite{jiang2020coherent} & \cellcolor{Gray}69.1 & \cellcolor{Gray}72.2 \\
3DCrowdNet {\scriptsize(Ours)} / {\scriptsize OpenPose}~\cite{cao2017realtime} & 70.2 & 70.9 \\
\cellcolor{Gray}\textbf{3DCrowdNet {\scriptsize(Ours)} / {\scriptsize HigherHRNet}~\cite{cheng2020higherhrnet}} & \cellcolor{Gray}\textbf{72.7} & \cellcolor{Gray}\textbf{73.3} \\
\specialrule{.1em}{.05em}{.05em}
\end{tabular}
\vspace*{-0.6em}
    \caption{Comparison on MuPoTS~\cite{mehta2018single} between 3DCrowdNet and previous methods. The numbers denote 3DPCK for all annotations (All) and annotations matched to a prediction (Matched), and are brought from~\cite{jiang2020coherent}. The method names beside~\cite{cao2017realtime,he2017mask,cheng2020higherhrnet} indicate the source of bounding boxes and 2D pose input. 
    }
\label{table:sota_mupots}
\vspace*{-0.5em}
\end{table}

\begin{figure*}[t]
\begin{center}
\includegraphics[width=0.88\linewidth]{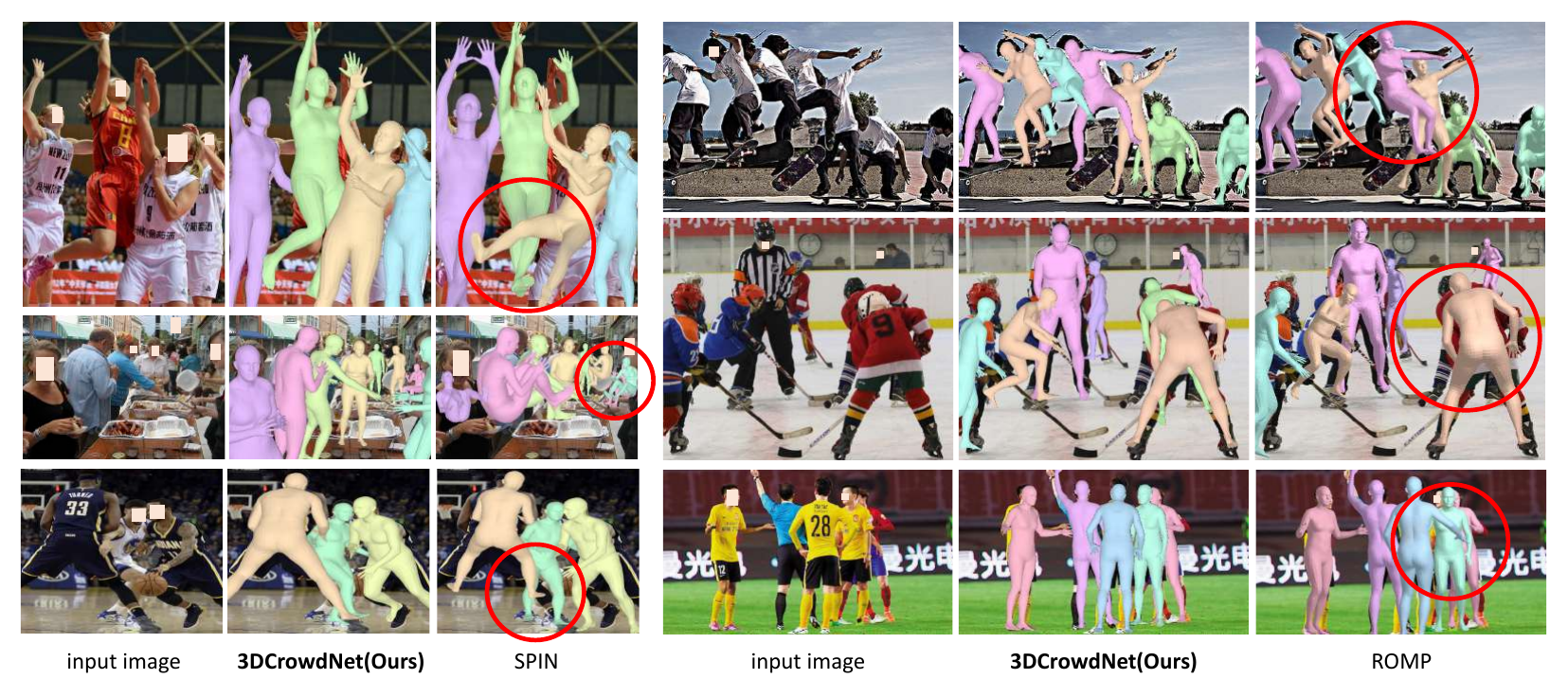}
\end{center}
\vspace*{-2em}
  \caption{
  Qualitative comparison on a CrowdPose~\cite{li2019crowdpose} test set with SPIN~\cite{kolotouros2019learning} and ROMP~\cite{sun2021monocular}. We highlighted their representative failure cases with red circles. The order of 3D meshes is manually assigned. 
  }
\label{fig:quality_comparison}
\vspace*{-0.8em}
\end{figure*}

\noindent\textbf{MuPoTS.}
Table~\ref{table:sota_mupots} compares our 3DCrowdNet with methods that recover a 3D mesh.
It outperforms all the previous methods.
Note that the second and fifth rows prove that 3DCrowdNet's high accuracy on the crowded scenes is not simply attributed to better localization derived from bottom-up 2D poses.
While 3DCrowdNet and HMR use the same 2D poses of OpenPose~\cite{cao2017realtime}, HMR utilizes the 2D pose only to get a bounding box, and 3DCrowdNet additionally uses the 2D pose to guide a feature extractor to extract crowded scene-robust image features.
Leveraging more information in given input is natural, and leads to better accuracy.

\begin{table}[t]
\small
\centering
\setlength\tabcolsep{1.0pt}
\def\arraystretch{1.1}
\begin{tabular}{C{3.4cm}|C{0.9cm}C{0.9cm}C{0.9cm}C{0.9cm}|C{0.9cm}}
\specialrule{.1em}{.05em}{.05em}
method & Haggl. & Mafia & Ultim. & Pizza & Mean \\ \hline
Zanfir~\etal~\cite{zanfir2018monocular} & 140.0 & 165.9 & 150.7 & 156.0 & 153.4 \\
\cellcolor{Gray}Zanfir~\etal~\cite{zanfir2018deep} & \cellcolor{Gray}141.4 & \cellcolor{Gray}152.3 & \cellcolor{Gray}145.0 & \cellcolor{Gray}162.5 & \cellcolor{Gray}150.3  \\
Jiang~\etal~\cite{jiang2020coherent} & 129.6 & 133.5 & 153.0 & 156.7 & 143.2 \\
\cellcolor{Gray}ROMP~\cite{sun2021monocular} & \cellcolor{Gray}111.8 & \cellcolor{Gray}\textbf{129.0} & \cellcolor{Gray}148.5 & \cellcolor{Gray}149.1 & \cellcolor{Gray}134.6 \\
REMIPS~\cite{fieraru2021remips} & 121.6 & 137.1 & 146.4 & 148.0 & 138.3 \\
\cellcolor{Gray}\textbf{3DCrowdNet {\scriptsize(Ours)}} & \cellcolor{Gray}\textbf{109.60} & \cellcolor{Gray}135.9 & \cellcolor{Gray}\textbf{129.8} & \cellcolor{Gray}\textbf{135.6} & \cellcolor{Gray}\textbf{127.6}\\

\specialrule{.1em}{.05em}{.05em}
\end{tabular}
\vspace*{-0.6em}
    \caption{Comparison on CMU-Panoptic~\cite{joo2017panoptic}. The numbers denote MPJPE. We follow the  evaluation protocol of Jiang~\etal~\cite{jiang2020coherent}. 
    }
\label{table:sota_cmu}
\vspace*{-1.0em}
\end{table}

\noindent\textbf{CMU-Panoptic.}
Table~\ref{table:sota_cmu} shows that our 3DCrowdNet significantly outperforms previous 3D human pose and shape estimation methods on CMU-Panoptic.
The result demonstrates that the proposed 3DCrowdNet can perform competitively on crowded scenes with daily social activities.
Note that no data from CMU-Panoptic are used for training.

\begin{table}[t]
\small
\centering
\setlength\tabcolsep{1.0pt}
\def\arraystretch{1.1}
\begin{tabular}{C{3.4cm}|C{1.3cm}C{2.0cm}C{1.3cm}}
\specialrule{.1em}{.05em}{.05em}
method & MPJPE$\downarrow$ & PA-MPJPE$\downarrow$ & MPVPE$\downarrow$ \\ \hline
HMR~\cite{kanazawa2018end} & 130 & 76.7 & -\\
\cellcolor{Gray}GraphCMR~\cite{kolotouros2019convolutional} & \cellcolor{Gray}- & \cellcolor{Gray}70.2 & \cellcolor{Gray}- \\
SPIN~\cite{kolotouros2019learning} & 96.9 & 59.2 & 116.4\\
\cellcolor{Gray}I2L-MeshNet~\cite{moon2020i2l} & \cellcolor{Gray}93.2 & \cellcolor{Gray}57.7 & \cellcolor{Gray}110.1 \\
Pose2Mesh~\cite{choi2020p2m} & 89.5 & 56.3 & 105.3 \\
\cellcolor{Gray}Song~\etal~\cite{song2020human} & \cellcolor{Gray}- & \cellcolor{Gray}55.9 & \cellcolor{Gray}- \\
Fang~\etal~\cite{fang2021reconstructing} & 85.1 & 54.8 & -\\
\cellcolor{Gray}TUCH~\cite{muller2021self} & \cellcolor{Gray}84.9 & \cellcolor{Gray}55.5 & \cellcolor{Gray}-\\
ROMP~\cite{sun2021monocular} & 91.3 & 54.9 & 108.3 \\
\cellcolor{Gray}\textbf{3DCrowdNet {\scriptsize(Ours)}} & \cellcolor{Gray}\textbf{81.7} & \cellcolor{Gray}\textbf{51.5} & \cellcolor{Gray}\textbf{98.3} \\
\specialrule{.1em}{.05em}{.05em}
\end{tabular}
\vspace*{-0.6em}
    \caption{Comparison on 3DPW~\cite{von20183dpw} between 3DCrowdNet and state-of-the-art methods of 3D human mesh estimation from a single image.
    We compare methods that do not use 3DPW train set during training for the fair comparison.
    }
\label{table:sota_3dpw}
\vspace*{-1em}
\end{table}

\noindent\textbf{3DPW.}
Table~\ref{table:sota_3dpw} shows that 3DCrowdNet achieves state-of-the art accuracy in general in-the-wild scenes.
The result validates that 3DCrowdNet is robust to diverse challenges of in-the-wild scenes, although our method is designed to target crowded scenes.
The second row of Figure~\ref{fig:quality_comparison} supports our statement, which shows 3DCrowdNet's robustness to truncation and occlusion in in-the-wild images.

We provide the qualitative comparison with SPIN~\cite{kolotouros2019learning} and ROMP~\cite{sun2021monocular} in Figure~\ref{fig:quality_comparison}.
Apparently, 3DCrowdNet produces much more robust 3D meshes on in-the-wild crowded scenes.
SPIN predicts a swapped leg pose (top), fails to distinguish different people in the overlapping bounding boxes (middle), and misses the right leg's pose due to inter-person occlusion (bottom).
ROMP produces an inaccurate pose for a person under occlusion with similar appearances (top), misses a target whose body center (\textit{i.e.} torso) is invisible (middle), and estimate an inaccurate global rotation of a target due to occlusion by a nearby person with similar appearance (bottom).
Please also refer to more extensive qualitative comparison with~\cite{kolotouros2019learning,moon2020i2l,choi2020p2m,sun2021monocular} and failure cases of 3DCrowdNet in the supplementary material.

\section{Conclusion}
We present 3DCrowdNet, the first single image-based 3D human mesh estimation system that explicitly targets in-the-wild crowded scenes. 
It extracts crowded-scene robust image features of a target person, and effectively distinguishes the target from others.
We guide a deep CNN to pay attention to the target using a 2D pose, which is robust to the domain gap between MoCap training data and crowd testing data.
The joint-based regressor preserves the spatial activation of the target, and effectively excludes non-target people's image features.
We show that 3DCrowdNet highly outperforms previous methods on in-the-wild crowded scenes both quantitatively and qualitatively.
3DCrowdNet could be a baseline for future image-based methods that target crowded scenes owing to the simple yet effective implementation.

\noindent\textbf{Acknowledgements.}
This work was supported in part by IITP grant funded by the Korea government (MSIT) [No. 2021-0-01343, Artificial Intelligence Graduate School Program (Seoul National University)].

{\small
\bibliographystyle{ieee_fullname}
\bibliography{main}
}

\clearpage

\twocolumn[{
\small
\begin{center}\centering
\setlength\tabcolsep{1.0pt}
\def\arraystretch{1.1}
\begin{tabular}{C{3.4cm}|C{8cm}|C{1.3cm}C{2.0cm}C{1.3cm}}
\specialrule{.1em}{.05em}{.05em}
method & training datasets & MPJPE$\downarrow$ & PA-MPJPE$\downarrow$ & MPVPE$\downarrow$ \\ \hline
SPIN~\cite{kolotouros2019learning} & Human3.6M~\cite{ionescu2014human3}, MPI-INF-3DHP~\cite{mehta2017monocular}, MSCOCO~\cite{lin2014mscoco}, MPII~\cite{andriluka2014mpii}, LSP~\cite{johnson2010lsp}, LSP-Extended~\cite{johnson2011lspex} & 121.2 & 69.9 & 144.1\\
\cellcolor{Gray}Pose2Mesh~\cite{choi2020p2m} & \cellcolor{Gray}Human3.6M~\cite{ionescu2014human3}, MuCo-3DHP~\cite{mehta2018single}, MSCOCO~\cite{lin2014mscoco} & \cellcolor{Gray}124.8 & \cellcolor{Gray}79.8 & \cellcolor{Gray}149.5 \\
I2L-MeshNet~\cite{moon2020i2l} & Human3.6M~\cite{ionescu2014human3}, MuCo-3DHP~\cite{mehta2018single}, MSCOCO~\cite{lin2014mscoco} & 115.7 & 73.5 & 162.0 \\
\cellcolor{Gray}ROMP~\cite{sun2021monocular} & \cellcolor{Gray}Human3.6M~\cite{ionescu2014human3}, MPI-INF-3DHP~\cite{mehta2017monocular}, MSCOCO~\cite{lin2014mscoco}, MPII~\cite{andriluka2014mpii}, LSP~\cite{johnson2010lsp}, LSP-Extended~\cite{johnson2011lspex}, AICH~\cite{wu2017ai}, MuCo-3DHP~\cite{mehta2018single}, OH~\cite{zhang2020object}, PoseTrack~\cite{andriluka2018posetrack}, CrowdPose~\cite{li2019crowdpose} & \cellcolor{Gray}104.8 & \cellcolor{Gray}63.9 & \cellcolor{Gray}127.8 \\
\textbf{3DCrowdNet (Ours)} & \textbf{MuCo-3DHP~\cite{mehta2018single}, MSCOCO~\cite{lin2014mscoco}} & \textbf{88.3} & \textbf{59.2} & \textbf{112.8} \\
\specialrule{.1em}{.05em}{.05em}
\end{tabular}
\captionof{table}{Comparison on 3DPW-Crowd between 3DCrowdNet and previous methods. 3DCrowdNet uses the least training datasets and achieves the best accuracy on in-the-wild crowded scenes.}
\label{table:sota_train_data}
\vspace*{2.0em}
\end{center}
}]

\noindent\textbf{\Large{Appendix}}
\section{Datasets}

\subsection{Training sets of different methods}
\label{supp:train_set}

Table~\ref{table:sota_train_data} demonstrates that the superiority of 3DCrowdNet does not come from using more training data.
It shows the training datasets used in the previous methods of the main manuscript's Table 5.
We trained 3DCrowdNet on one MoCap dataset and one in-the-wild 2D dataset, which is the least training set among methods, and we tested it on 3DPW-Crowd.
It still significantly outperforms the previous methods in all metrics.
We used 2D pose outputs of HigherHRNet~\cite{cheng2020higherhrnet}, which is trained only on MSCOCO~\cite{lin2014mscoco}.
The results strongly support that our contributions listed in the main manuscript's Section 1.

\subsection{Details of testing sets}
\label{supp:test_set}
\noindent\textbf{3DPW-Crowd.}
The sequence names of 3DPW-Crowd are \textit{courtyard\_hug\_00} and \textit{courtyard\_dancing\_00}, a subset of the 3DPW~\cite{von20183dpw} validation set.
3DPW-Crowd contains 1073 images and 1923 persons with GT 3D pose and shape annotations.
The average bounding box IoU is 37.5\%, and the CrowdIndex~\cite{li2019crowdpose} is 49.3\%.
We used 14 joints defined by Human3.6M~\cite{ionescu2014human3} for evaluating PA-MPJPE and MPJPE following the previous works~\cite{kolotouros2019learning,choi2020p2m,moon2020i2l}.


\noindent\textbf{MuPoTS.}
MuPoTS~\cite{mehta2018single} contains 20 sequences, 8370 images, and 20899 persons with GT 3D pose annotations.
The sequences are captured indoors and outdoors, and GT 3D poses are obtained by a multi-view marker-less motion capture system.
The average bounding box IoU is 3.8\%, and the CrowdIndex~\cite{li2019crowdpose} is 13.2\%.
We used the official MATLAB code for evaluation.

\noindent\textbf{CMU-Panoptic.}
We selected four sequences that show people doing social activities, namely \textit{Haggling}, \textit{Mafia}, \textit{Ultimatum}, and \textit{Pizza} following~\cite{zanfir2018monocular,jiang2020coherent}.
Sequences captured by the 16th and 30th cameras are selected.
The sequences contain 9600 frames and 21,404 persons with GT 3D pose annotations.
The average bounding box IoU is 2.0\%, and the CrowdIndex~\cite{li2019crowdpose} is 11.1\%.
We used pre-processed GT annotations and followed the evaluation protocol of~\cite{jiang2020coherent} in their official code repository.

\noindent\textbf{3DPW.}
We used the test set of 3DPW~\cite{von20183dpw} following the official split protocol.
The test set contains 26240 images and 35515 persons with GT 3D pose and shape annotations.
The average bounding box IoU is 3.7\%, and the CrowdIndex~\cite{li2019crowdpose} is 4.9\%.
Sequences starring one actor are excluded in computing the bounding box IoU and the CrowdIndex.
We used 14 joints defined by Human3.6M~\cite{ionescu2014human3} for evaluating PA-MPJPE and MPJPE following the previous works~\cite{kolotouros2019learning,choi2020p2m,moon2020i2l}.
\section{More qualitative results}
\label{supp:quality}

\begin{figure}[!t]
\begin{center}
\includegraphics[width=1.0\linewidth]{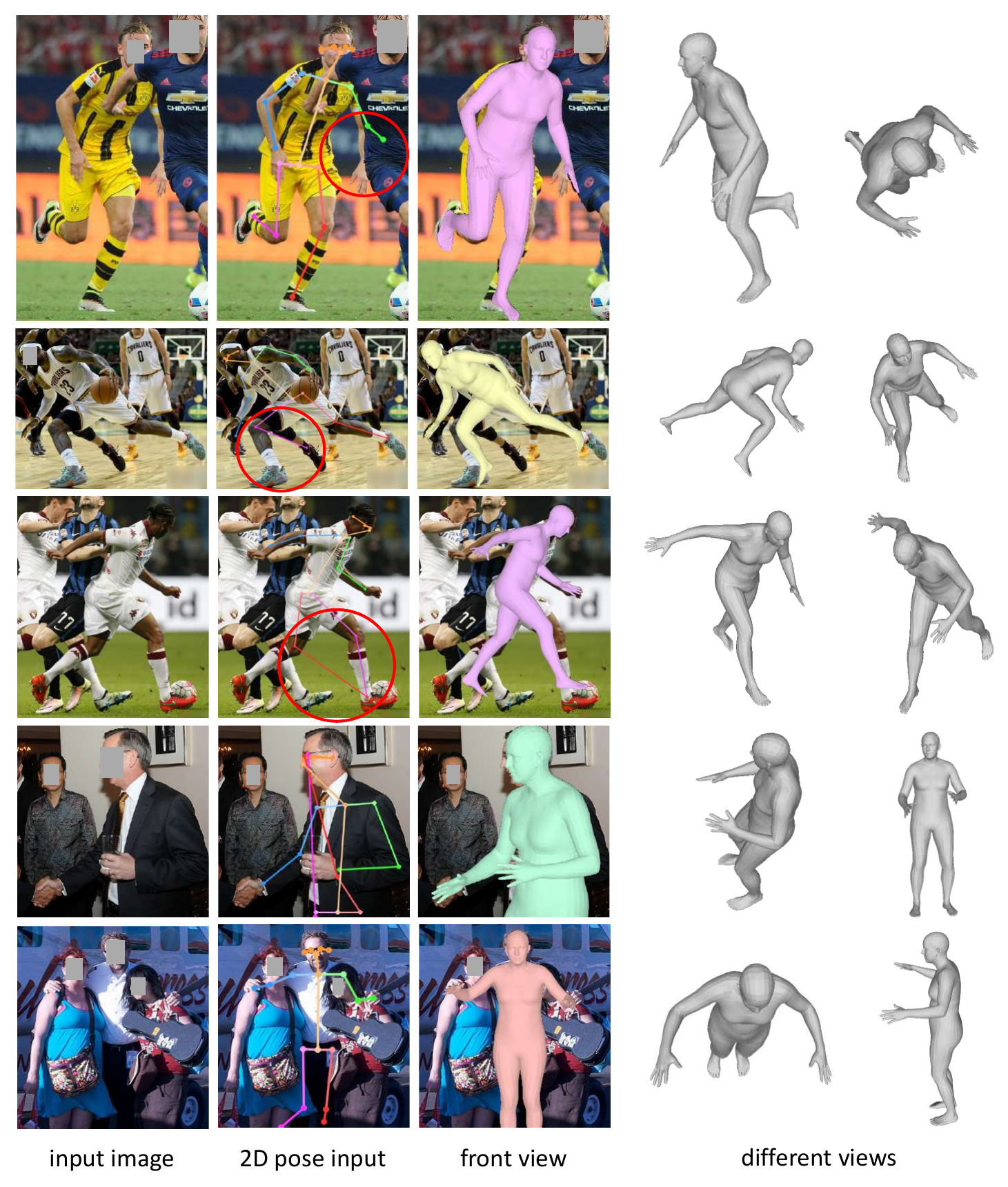}   
\end{center}
  \caption{
  Visualization of 3D meshes from different viewpoints. Our 3DCrowdNet can recover a 3D shape that best describes the target person in an image, even when provided with inaccurate 2D poses, using the target person's image features.
  }
\label{fig:pose_correction}
\end{figure}

\begin{figure}[!t]
\begin{center}
\includegraphics[width=1.0\linewidth]{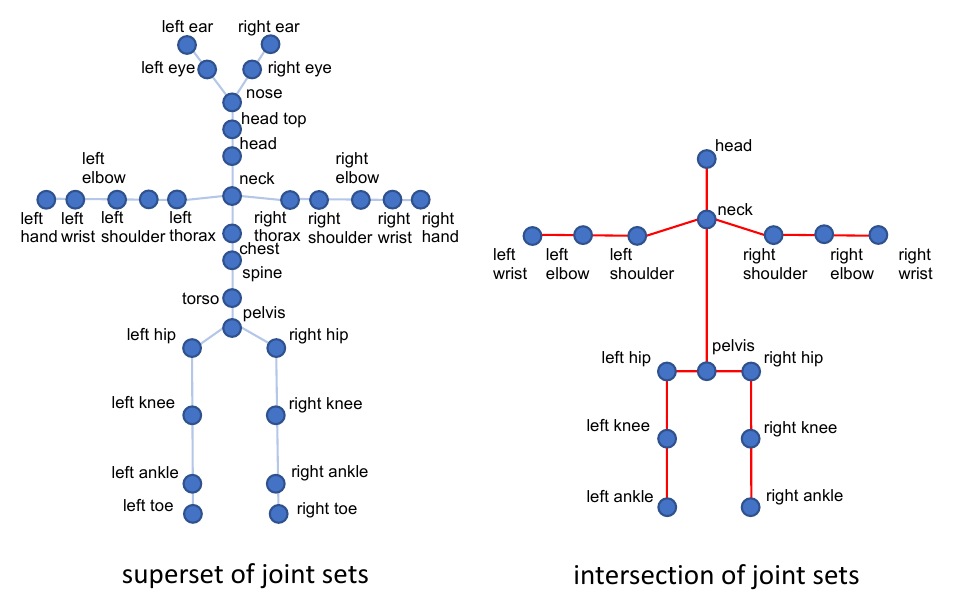}
\end{center}
  \caption{
  Illustration of joint sets. The red skeleton of the intersection of joint sets defines the joints' neighborhood of graph convolution used in the joint-based regressor. 
  }
\label{fig:joint_sets}
\end{figure}

\noindent\textbf{Accurate 3D meshes from erroneous 2D pose input.}
Figure~\ref{fig:pose_correction} shows that our 3DCrowdNet can estimate robust 3D meshes, given inaccurate 2D poses from in-the-wild crowded scenes.
Due to inter-person occlusion and overlapping bounding boxes between people, 2D pose estimators~\cite{cao2017realtime,cheng2020higherhrnet} may produce inaccurate joint predictions as shown in the first, second, and third rows.
To handle such cases, the feature extractor of 3DCrowdNet assigns don't-care values to the joint predictions with low confidence (\textit{e.g.} lower than $0.1$ for outputs from~\cite{cheng2020higherhrnet}) using heatmap representation, as discussed in Section 3.1 of the main manuscript.
Then, the joint-based regressor of 3DCrowdNet refines the 2D pose heatmap, while predicting a 3D pose with image features containing the image's context information.
The 2D pose heatmap has a different joint set, a superset of joint sets defined by multiple datasets, with the 3D pose's joint set, an intersection of joint sets defined by multiple datasets.
Figure~\ref{fig:joint_sets} depicts each joint set.
Last, the joint-based regressor samples image features using the ($x$,$y$) pixel positions of the 3D pose and estimates human model parameters, SMPL~\cite{loper2015smpl} parameters.
The joint-based regressor's graph convolutional layers refines the image features of joint predictions by fully exploiting the human kinematic prior and regress parameters of a 3D mesh that best describes a target person in a crowd.
The fourth and fifth rows of Figure~\ref{fig:pose_correction} prove that our approach is also effective on estimating robust 3D meshes from truncated images, which often have missing 2D joint predictions.

\begin{figure*}
\begin{center}
\includegraphics[width=0.9\linewidth]{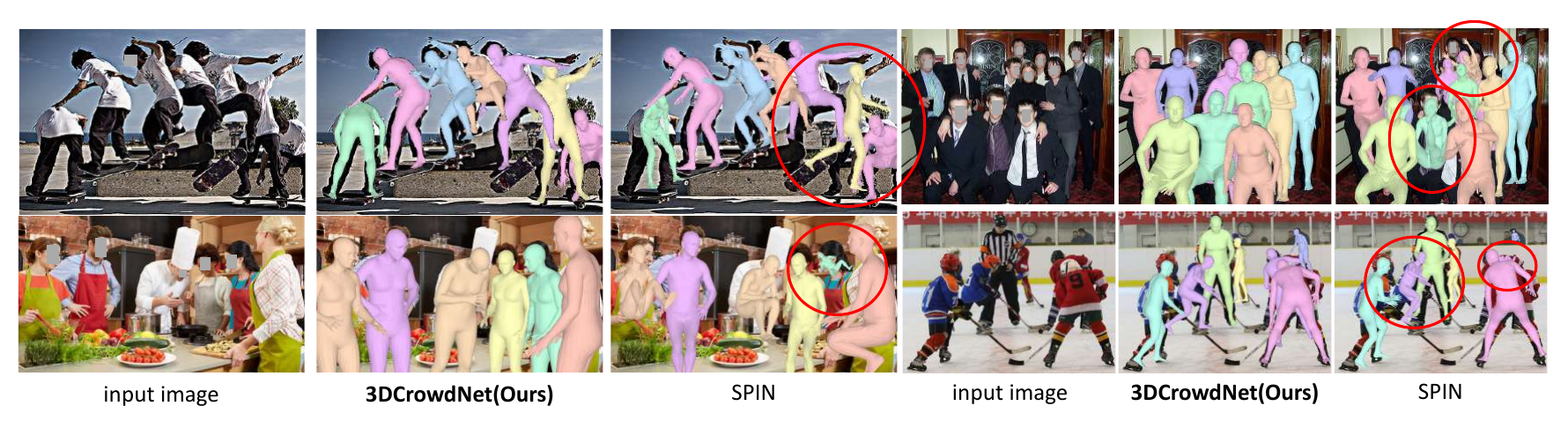}\end{center}
\vspace*{-3mm}
  \caption{
  Qualitative comparison on the CrowdPose~\cite{li2019crowdpose} test set.  We highlighted the failure cases of SPIN~\cite{kolotouros2019learning} with red circles. SPIN tends to be sensitive to occlusion, while 3DCrowdNet provides robust 3D meshes.
  }
\label{fig:spin_comparison}
\end{figure*}

\begin{figure*}
\begin{center}
\includegraphics[width=0.9\linewidth]{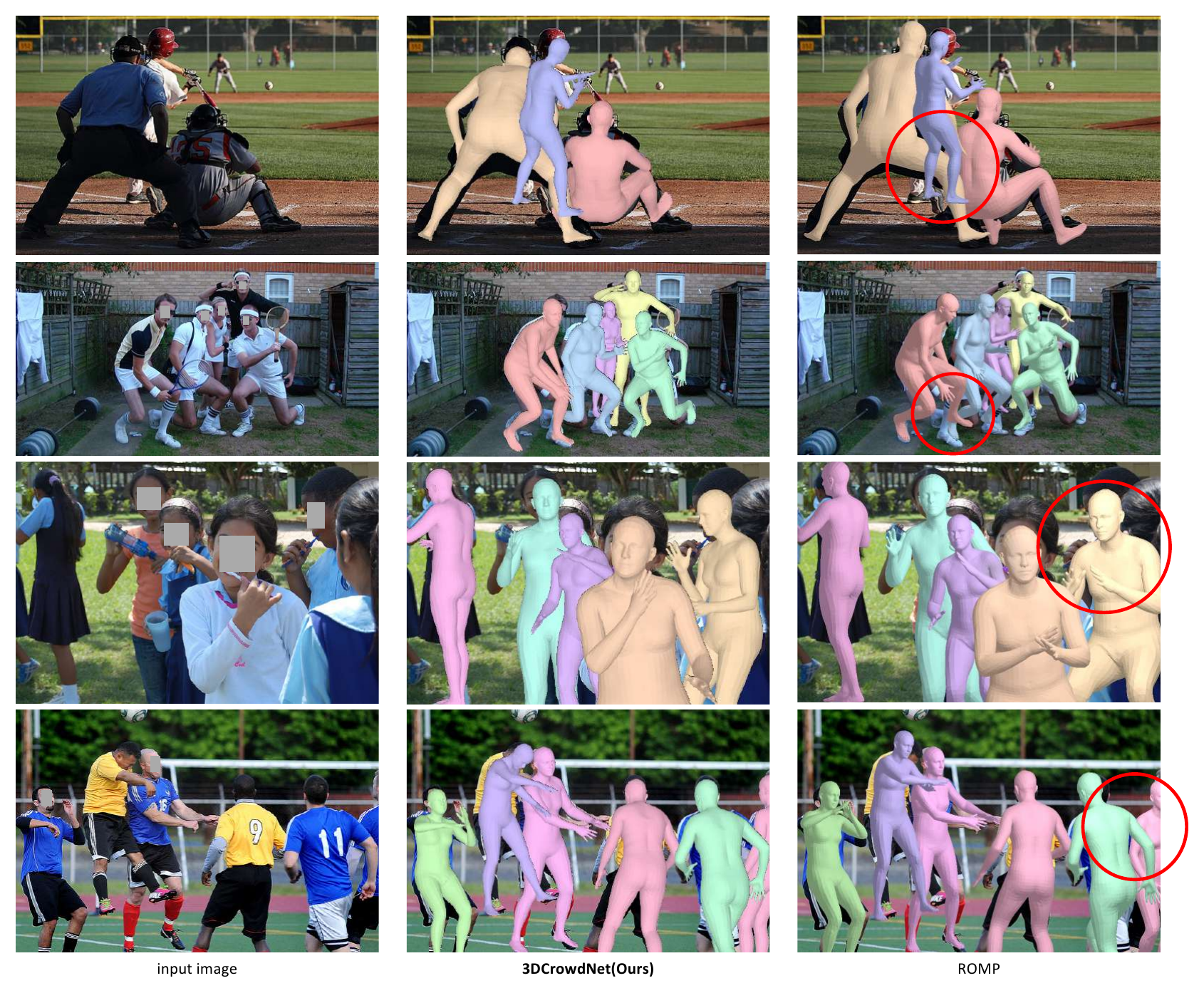}
\end{center}
\vspace*{-5mm}
  \caption{
  Qualitative comparison on the CrowdPose~\cite{li2019crowdpose} test set. We highlighted the failure cases of ROMP~\cite{sun2021monocular} with red circles. Wrong global rotation of occluded persons (the third and fourth rows); inaccurate leg poses under inter-person occlusion (the first and third rows). 3DCrowdNet produces much more robust 3D meshes.
  }
\vspace*{-3mm}
\label{fig:romp_comparison}
\end{figure*}

\begin{figure*}
\vspace*{2mm}
\begin{center}
\includegraphics[width=1.0\linewidth]{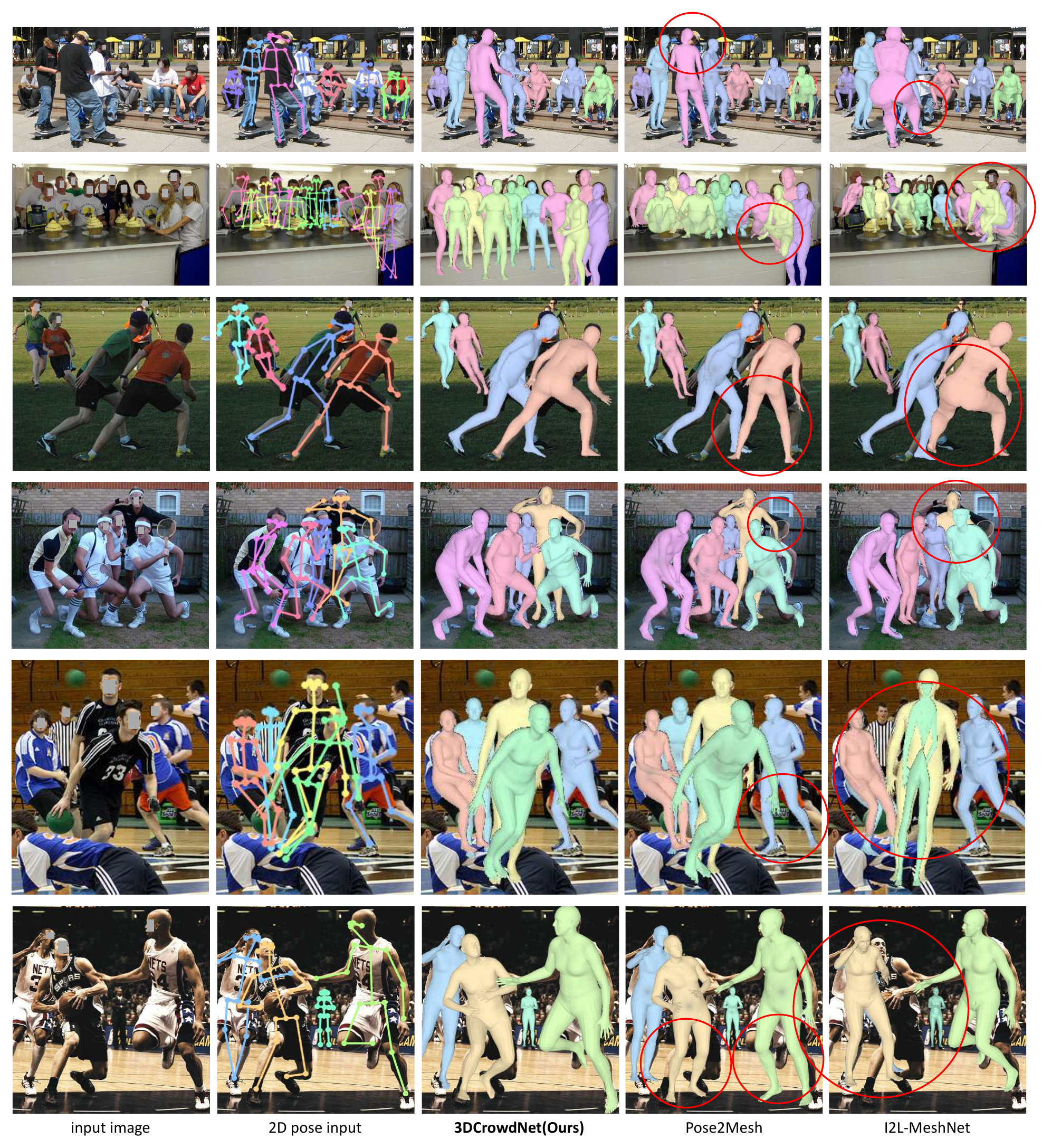}
\end{center}
\vspace*{-3mm}
  \caption{
  Qualitative comparison on the CrowdPose~\cite{li2019crowdpose} test set. From left, an input image, 2D pose input, 3DCrowdNet, I2L-MeshNet~\cite{moon2020i2l}, and Pose2Mesh~\cite{choi2020p2m} outputs. Our 3DCrowdNet successfully disentangles a target person from other people in a bounding box compared with I2L-MeshNet. Also, 3DCrowdNet produces a 3D shape that best describes a target person in images, while Pose2Mesh estimates a plausible 3D shape for given 2D poses, which does not correspond to input images. 3DCrowdNet and I2L-MeshNet use the same bounding boxes to crop an image for each person. 3DCrowdNet and Pose2Mesh use the same 2D poses from ~\cite{cheng2020higherhrnet}.
  }
\vspace*{-3mm}
\label{fig:i2l_p2m_quality_comparison}
\end{figure*}

\noindent\textbf{Comparison with SPIN.}
We provide more qualitative comparison with SPIN~\cite{kolotouros2019learning} in Figure~\ref{fig:spin_comparison}.
SPIN is one of the state-of-the-art methods that are based on the two wheels of the current 3D human mesh estimation literature, the mixed batch training and the model-based approach using a global feature discussed in Section $1$ of the main manuscript.
Our 3DCrowdNet produces accurate and robust 3D meshes from diverse in-the-wild crowded scenes.
On the other hand, SPIN predicts an incorrect overall pose for a person under severe inter-person occlusion (top-left), estimates inaccurate leg poses (bottom-left, bottom-right), and produces noisy 3D meshes (top-right, bottom-left, bottom-right) that show vulnerability to inter-person occlusion. 

\begin{figure*}
\begin{center}
\includegraphics[width=1.0\linewidth]{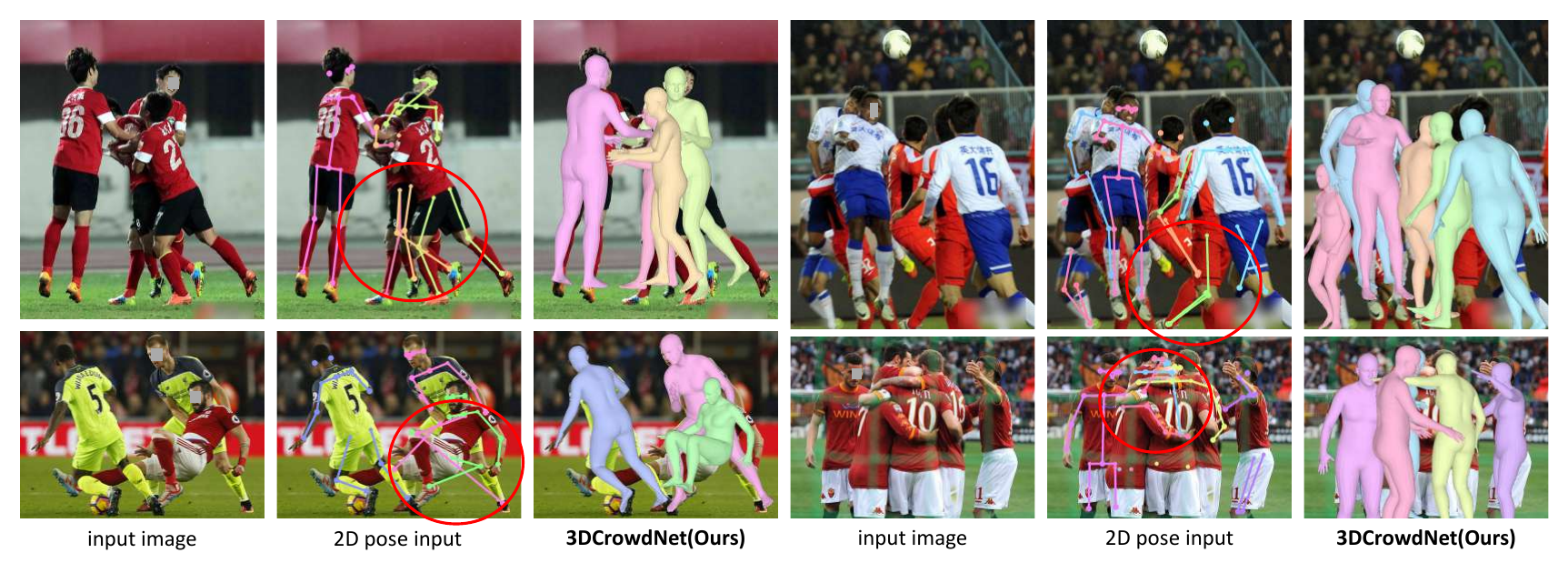}
\end{center}
\vspace*{-3mm}
  \caption{
  Failure cases of 3DCrowdNet.
  }
\label{fig:failure_quality}
\end{figure*}

\begin{figure}
\begin{center}
\includegraphics[width=1.0\linewidth]{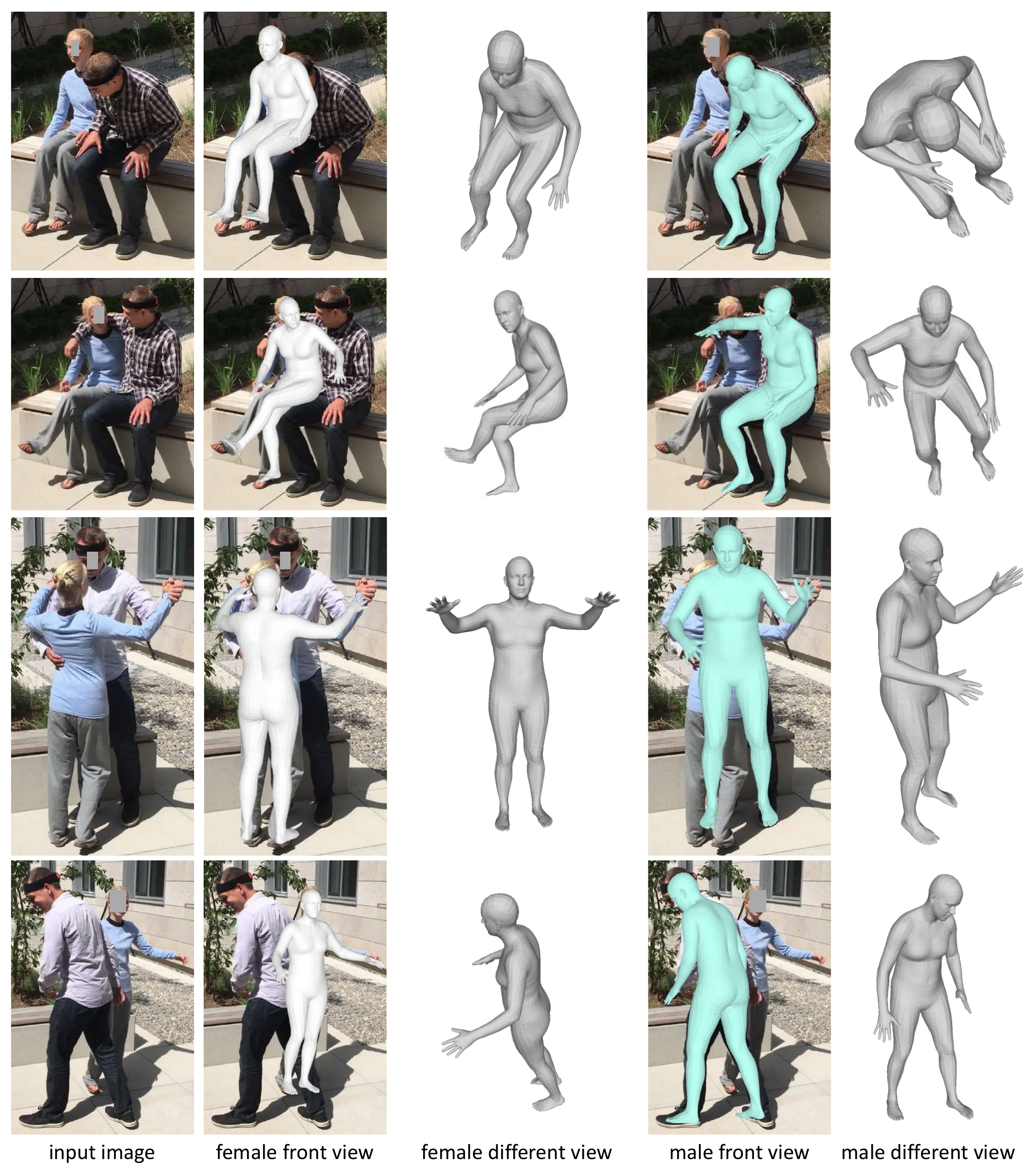}
\end{center}
  \caption{
 3DCrowdNet's outputs on 3DPW-Crowd.
  }
\label{fig:3dpw_crowd_result}
\end{figure}

\noindent\textbf{Comparison with ROMP.}
We provide more qualitative comparison with ROMP~\cite{sun2021monocular} in Figure~\ref{fig:romp_comparison}.
ROMP is a bottom-up method for multi-person 3D mesh estimation.
Our 3DCrowdNet produces accurate and robust 3D meshes from diverse in-the-wild crowded scenes.
On the contrary, ROMP predicts the wrong global rotation of a target person from in-the-wild crowded scenes (the third and fourth rows) and produces inaccurate leg poses under severe inter-person occlusion and crossed human parts (the first and second rows).

\noindent\textbf{Comparison with Pose2Mesh and I2L-MeshNet.}
Figure~\ref{fig:i2l_p2m_quality_comparison} shows the qualitative comparison between 3DCrowdNet, Pose2Mesh~\cite{choi2020p2m}, and I2L-MeshNet~\cite{moon2020i2l}.
Pose2Mesh and I2L-MeshNet are state-of-the-art model-free 3D mesh estimators, which predict coordinates of mesh vertices.
Especially, Pose2Mesh is one of the most relevant competitors, since it can also benefit from the same 2D pose input.
3DCrowdNet produces much more robust 3D meshes from in-the-wild crowded scenes than the two methods.
Pose2Mesh estimates the most plausible 3D mesh for a given 2D pose (the first and fourth rows), not the 3D mesh that best describes a target in a crowd, as discussed in Section $5$ of the main manuscript.
Also, it often wrongly corrects the 2D pose input and produces common standing leg poses different from the images (the third, fifth, and sixth rows).
I2L-MeshNet fails to distinguish different people in overlapping bounding boxes (the fifth and sixth rows).
In addition, it tends to provide very noisy 3D pose and shape of a target in crowded scenes, which reveals the method's vulnerability to inter-person occlusion.
The results in the second row also validate the superiority of 3DCrowdNet's robustness to truncated bodies.

\noindent\textbf{Results on 3DPW-Crowd.}
Figure~\ref{fig:3dpw_crowd_result} illustrates the 3DCrowdNet's outputs on 3DPW-Crowd.
3DCrowdNet estimates robust 3D pose and shape on images that show people having highly close interaction.
Different people in overlapping bounding boxes are disentangled, and occluded body parts are reasonably reconstructed.

\section{Limitation}
\label{supp:limit}

Although the proposed 3DCrowdNet highly outperforms the previous 3D mesh estimation methods in in-the-wild crowded scenes, there is a limitation to be resolved in future work.
As shown in Figure~\ref{fig:failure_quality}, when the 2D pose is inaccurate and appearances of nearby persons are very similar, 3DCrowdNet fails to produce robust 3D meshes.
The top-left and bottom-right cases of Figure~\ref{fig:failure_quality} are the representative cases, which can be easily found in sports images.
In such cases, it is challenging for 3DCrowdNet to correct the inaccurate 2D pose with image features, since the context information in image features is ambiguous due to indistinguishable appearances.
One way of resolving the challenge could be to model the relative translation between persons to better understand the context.
Alternatively, data augmentation to make a network robust to similar appearances would be an interesting direction.

\section{2D pose estimators.}
\label{supp:2dpose}

In this work, we used 2D pose outputs from OpenPose~\cite{cao2017realtime} and HigherHRNet~\cite{cheng2020higherhrnet}.
The OpenPose outputs used in 3DPW~\cite{von20183dpw} are included in the annotations of 3DPW~\cite{von20183dpw}.
The OpenPose used in MuPoTS~\cite{mehta2018single} are obtained by running the third-party PyTorch~\cite{paszke2017automatic} code implementation\footnote{\url{https://github.com/tensorboy/pytorch\_Realtime\_Multi-Person\_Pose\_Estimation}}.
OpenPose is trained on \textbf{COCO2017 train}~\cite{lin2014mscoco} dataset.
It achieves 65.3 mAP (mean Average Precision) in \textbf{COCO2017 val} dataset.
In the CrowdPose~\cite{li2019crowdpose} test set, it achieves 48.7 and 32.3 mAPs for medium and hard cases, respectively.
All the HigherHRNet outputs are obtained by running the official code implementation.
HigherHRNet is trained on \textbf{COCO2017 train} dataset.
It achieves 0.671 mAP on \textbf{COCO2017 val} dataset.
In the CrowdPose~\cite{li2019crowdpose} test set, it achieves 68.1 and 58.9 mAPs for medium and hard cases, respectively.

\subsection*{License of the Used Assets}
\begin{compactitem}[$\bullet$]
    \item MSCOCO dataset~\cite{lin2014mscoco} belongs to the COCO Consortium and are licensed under a Creative Commons Attribution 4.0 License.
    \item Human3.6M dataset~\cite{ionescu2014human3}'s licenses are limited to academic use only. 
    \item MPII dataset~\cite{andriluka2014mpii} is released for academic research only and it is free to researchers from educational or research institutes for non-commercial purposes.
    \item 3DPW dataset~\cite{von20183dpw} is released for academic research only and it is free to researchers from educational or research institutes for non-commercial purposes.
    \item CrowdPose dataset~\cite{li2019crowdpose} is released for academic research only and it is free to researchers from educational or research institutes for non-commercial purposes.
    \item MuCo-3DHP and MuPoTS~\cite{mehta2018single} are released for any non-commercial purposes.
    \item CMU-Panoptic~\cite{joo2017panoptic} is released only for research purposes.
    \item The third party implementation\footnote{\url{https://github.com/tensorboy/pytorch\_Realtime\_Multi-Person\_Pose\_Estimation}} of OpenPose~\cite{cao2017realtime} is licensed under the MIT license.
    \item HigherHRNet~\cite{cheng2020higherhrnet}'s implementation\footnote{\url{https://github.com/HRNet/HigherHRNet-Human-Pose-Estimation}} is licensed under the MIT license.
\end{compactitem}

\end{document}